# A Latent Variable Approach for
# Non-Hierarchical Multi-Fidelity Adaptive Sampling

*Yi-Ping Chen[1], Liwei Wang[1], Yigitcan Comlek[1], Wei Chen[1]\**
*[1]Department of Mechanical Engineering, Northwestern University, Evanston, IL, USA*

## Abstract

Multi-fidelity (MF) methods are gaining popularity for enhancing surrogate modeling and design optimization by incorporating data from both high- and various low-fidelity (LF) models. While most existing MF methods assume a fixed training set, adaptive sampling methods that dynamically allocate resources among models with different fidelities can achieve higher efficiency in the exploration and exploitation of the design space. However, these methods either rely on the hierarchical assumption of fidelity levels or fail to capture their intercorrelation which is critical in quantifying the benefit of future samples for the adaptive sampling. To address this hurdle, we propose an MF adaptive sampling framework hinged on a latent embedding for different fidelity models and an associated pre-posterior analysis to explicitly utilize their correlations to quantify the benefit of the candidate samples as the sampling criteria. In this framework, each infill sampling iteration includes two steps: First, we identify the HF location of interest with the greatest potential improvement of the high-fidelity (HF) model, and then search for the next sample across all fidelity levels that maximizes the improvement per unit cost at the location identified in the first step. This is made possible by a single Latent Variable Gaussian Process (LVGP) model that maps different fidelity models into an interpretable latent space to capture their correlations without assuming any hierarchy between fidelity levels. The LVGP enables us to assess how LF sampling candidates will affect HF response with a pre-posterior analysis and determine the next sample with the best benefit-to-cost ratio. Furthermore, the proposed method offers the flexibility to switch between global fitting (GF) and Bayesian Optimization (BO) by simply changing the acquisition function. Through test cases, we demonstrate that our method outperforms state-of-the-art methods in both MF GF and BO problems in the rate of convergence and robustness.

*\*Corresponding Author: Professor Wei Chen, weichen@northwestern.edu; 2145 Sheridan Rd., Evanston, IL, 60208*

## 1. Introduction

Multi-fidelity (MF) methods have been gaining more attention in recent years, emphasizing the need for combining data from high-fidelity (HF) and low-fidelity (LF) models [1–3]. An HF model can precisely approximate the real and detailed behavior of a system but may be expensive for simulation or data acquisition. LF models, on the other hand, approximate a system in a simplified or less detailed way, such as simplified analytical models [4,5], reduced order models (ROM) [6,7], finite element models with coarse meshes [8,9], or data-driven machine learning models [1,10,11]. MF methods can also be associated with the use of different physical laws [12,13], e.g. linear and nonlinear constitutive assumptions [14]. In digital twin systems [15], MF methods can be applied for off-line data integration (e.g. fusing simulation and experimental data [16–18]) and online data fusion (e.g., fusing sensor data, surrogate model, HF model, etc. [19,20]). By integrating LF approximations with the HF model as an MF model, reasonably accurate results can be obtained with significantly reduced costs. Despite its promise, MF methods often rely on a high-quality training dataset, usually from HF data sources, to provide enough information for the learning process. To enhance adaptivity and efficiency, it is advantageous to enable active sampling for MF models, allowing them to actively select the most informative data points from low-cost data sources. The goal of this study is thus to develop such an MF adaptive sampling framework that integrates/fuses data from multiple fidelities and utilizes the learned correlations to realize cost-effective adaptive sampling, for either global fitting (GF) [21] or Bayesian Optimization (BO) [22,23].

In general, MF methods can be divided into hierarchical and non-hierarchical architecture based on how different data resources correlate in the modeling. By hierarchical architecture, we mean the fidelity levels of the simulation models need to be ranked in advance based on prior knowledge, and the MF surrogate models are trained sequentially by incorporating data from the lowest fidelity model to the highest one, as shown in Figure 1(a). Inspired by the Kennedy and O'Hagan (KOH) framework that captures model discrepancy with Gaussian Process (GP) models [24], a variety of GP-based MF methods have been widely developed for hierarchical structures. Some works assume linear relationships between two consecutive fidelity models [24,25], while others assume nonlinear relations to capture more complex correlations [26–29], and are extended to address problems where the HF and the LF models are different in design/input space [30] or input dimensions [31]. Another branch of hierarchical MF methods utilizes deep neural networks (DNNs), with an emphasis on engineering problems with higher dimensions and larger datasets [32–37]. The unique advantage of GP-based MF methods is the built-in uncertainty quantification, which allows adaptive sampling for GF [38,39], BO [2,40–43], and robust design optimization [44,45]. In contrast, DNN-based methods typically do not have inherent uncertainty quantification, so their integration with adaptive sampling typically uses entropy search-based methods [46–49] to quantify the benefit of infill samples and navigate the sampling strategies [43].

In practical engineering applications, these hierarchical MF methods exhibit significant limitations[30]. On the one hand, it is usually unclear how to rank the fidelity levels of the different models. For instance, multiple models developed for the same system may be built on different physics or mechanisms, for which the fidelity levels of various models may seem similar or unknown [31]. On the other hand, the hierarchical assumption not only oversimplifies and restricts the information passing among different fidelities but also leads to undesirable uncertainty accumulation from LF models to the HF model [50,51]. These limitations may undermine the accuracy of the HF predictions and lower the efficiency and flexibility of adaptive sampling strategies that rely on the predictive uncertainty of the fidelity models. Therefore, there is a notable interest in developing more flexible MF techniques to integrate information from multiple models when there is no clear ranking of model fidelities.

To address the aforementioned limitations, MF methods with non-hierarchical architectures have been proposed to bypass the hierarchical requirement of fidelity level. Instead of learning from the fidelity models sequentially, MF methods with non-hierarchical architectures learn the correlation between fidelities simultaneously, so that HF prediction essentially conditions on all training samples, see Figure 1(b) and Figure 1(c). One example is the multi-output GP using the linear model of coregionalization (LMC) where the outputs are expressed as linear combinations of independent functions [26,52]. In applications of LMC tailored for MF, each fidelity model is considered as one of the outputs, and the LMC is used to capture the correlation between them [43,53]. However, the drawback of this method is that it assumes equal contributions from different models while training the GP. This can be an issue in many MF applications, where certain sources or fidelity levels of data should be more informative than others [26].

To handle this issue, there is an emerging method that uses latent embedding to represent non-hierarchical correlation for MF problems. Some works used an additional latent dimension to represent the closeness and the hierarchy between the LF and the HF models, showing that all the samples collected from every fidelity model can be surrogated with a single model in which the information of the fidelity levels is represented on an additional latent dimension [54]. However, in these frameworks, knowledge of the fidelity level is still required before fitting the surrogate, which makes them inapplicable when the correlations between fidelity models need to be identified. Another approach is to use the Latent Map Gaussian Process (LMGP), a GP-based method that accommodates qualitative and quantitative variables [55]. It constructs the latent space that learns the correlation between various sources directly from the data rather than predefining the fidelity ranking. Via LMGP, one can visualize the correlation between fidelity models on the latent space, providing insight into the correlation directly from the training set. Due to the advantage of the learned correlation without any hierarchical assumption or knowledge of the fidelity ranking, we considered it as a suitable MF representation for developing adaptive sampling methods.

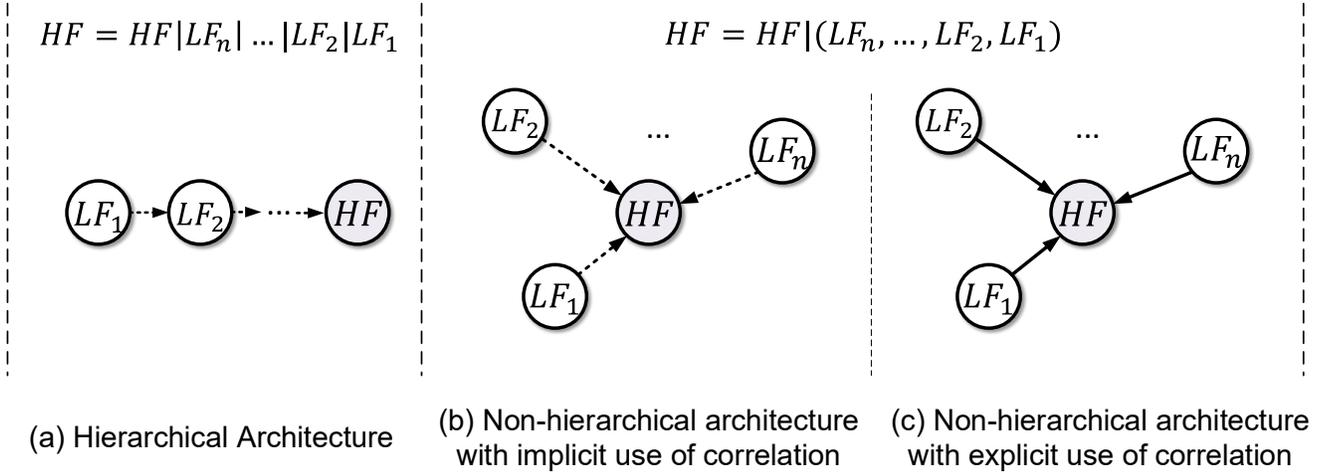

(a) Hierarchical Architecture

(b) Non-hierarchical architecture with implicit use of correlation

(c) Non-hierarchical architecture with explicit use of correlation

Figure 1: Different architectures of MF methods. (a) The hierarchical architecture where the HF model is recursively conditioned on the lower fidelity model. (b) The non-hierarchical architecture with implicit use of correlation. The dash arrows indicate that the correlations are only used for HF prediction and uncertainty quantification. The interaction between future LF infill samples and the HF response is neglected during the sampling process. (c) The non-hierarchical architecture of with explicit use of correlation. The solid arrows represent the information of the future benefit quantification of LF infill samples sharing between fidelity models to navigate the sampling strategy.

To determine *which location* and *which fidelity source* for infilling future samples, some methods *implicitly* utilize the correlation for response prediction while others *explicitly* use the correlation to quantify the impact of the infill samples to the HF performance. By *implicit* we mean that the correlation is only used to provide response and uncertainty predictions required by the acquisition function (AF). By qualitatively assuming that the improvement of LF models can benefit the HF models without any quantitative measure, the fidelity models are treated as separate sources and are assigned individual AFs. However, the interaction between LF samples and HF performance is neglected, i.e., how LF infill samples will impact HF response is unknown [10,41], as illustrated in Figure 1(b). Thus, these methods may lower the efficiency of adaptive sampling because the objective and the corresponding AF are designed to improve the LF models only but not explicitly consider the HF models. Moreover, it may result in the wrong convergence when the LF models are biased, i.e. it has a different optimum from the HF model [41].

In contrast, the *explicit* use of correlation is to quantify the benefit of the future samples by treating the MF system as a single surrogate model, and quantitatively capturing the influence of LF samples on HF response, as illustrated in Figure 1(c). By benefit, we mean how much improvement a certain sample can provide for the performance of the HF prediction. This improvement could manifest as the reduction of uncertainty in GF or the reduction in objective values (or acquisition) in BO. The effectiveness of an active sampling strategy relies on having a proper measure of this benefit and employing suitable inference methods to accurately assess its worth. At the core of this benefit quantification lies the challenge of quantifying how will the uncertainty/acquisition change in the region of interest by taking a future sample in any fidelity levels. It can only be realized if the uncertainty/acquisition information can be shared and approximated across fidelity models via the learned correlation. Therefore, by the explicit use of the correlation, the objective of the infill sampling would focus on the improvement of the HF model, maximizing the sampling efficiency.

We further categorize the approach to future benefit quantification into *point-to-model* and *point-to-point* quantification. The former approximates the aggregated change of uncertainty/acquisition of *the whole HF surrogate model* while evaluating the value of the candidate infill sample. It can be realized via the Information Gain [46], and can be included in the infill sampling strategies [48,50,56]. The latter, on the other hand, approximates the uncertainty/acquisition change at *any arbitrary location on the HF input space* before the infill sample is queried. It can be achieved by the pre-posterior analysis [57] that approximates the posterior distribution of the GP if a certain sample is added in the future. Note that the pre-posterior analysis can also perform point-to-model quantification by integrating the uncertainty of the approximated posterior distribution throughout the whole HF input space. Compared to the point-to-model quantification, the point-to-point quantification is expected to provide a better future benefit quantification, especially in BO because it focuses on improving the potential optimum only instead of that of the whole model. Thus, to maximize the flexibility of the proposed framework, we employ the pre-posterior analysis for future benefit quantification.

In this work, we propose a unified adaptive sampling framework, <u>Mu</u>lti-<u>F</u>idelity <u>A</u>daptive <u>Sa</u>mpling (MuFASa), that can perform both GF and BO compatible with most existing single-fidelity acquisition functions and any number of LF models. This framework comprises two pivotal elements: (a) we utilize the Latent Variable Gaussian Process (LVGP) to build a non-hierarchical MF surrogate, which falls into the latent representation methods and can be considered a generalized model of LMGP, noted as MF-LVGP, and (b) we apply the pre-posterior analysis on LVGP to assess the benefit of the infill samples on different fidelity models. By explicitly harnessing correlation information encoded in the latent space for pre-posterior analysis, our proposed framework accurately quantifies the benefit of future samples and guides the sampling process. Compared with existing methods, MuFASa exhibits the following advantages:

- It can be trained without requiring any prior knowledge of the hierarchy among fidelity levels and can accommodate non-hierarchical scenarios.
- It inherits the advantages of fast training and uncertainty quantification of GP.
- It embeds different fidelity levels into an interpretable latent space to capture their correlation and relationships.
- It can fully exploit the information from all the data in different fidelity levels, even from biased LF models, which is more data-efficient than existing methods that require relatively good LF models or discarding data.

- It exhibits greater efficiency, robustness, and sampling optimality for both GF and BO.

The rest of the paper is structured as follows. We introduce the mathematical formulation of MF-LVGP and the pre-posterior analysis in Section 2. The implementation details of MuFASa are revealed in Section 3. In Section 4. we use case studies to validate the method and compare it with the state-of-the-art to demonstrate its advantages. Finally, we conclude the paper in Section 5.

## 2. Multi-Fidelity Latent Variable Gaussian Process (MF-LVGP)

In this section, we introduce LVGP and its application to MF data fusion, noted as MF-LVGP. We further interpret and provide insights on the latent representation of different fidelity levels learned by MF-LVGP. Based on the insights, we elaborate on the pre-posterior analysis of LVGP.

### 2.1 Latent Variable Gaussian Process: Fitting and Predicting

Consider a modeling space with $\boldsymbol{w} = [\boldsymbol{x}^T, \boldsymbol{t}^T]^T$ where $\boldsymbol{x} = [x_1, x_2, \dots, x_q]^T \in R^q$ are the quantitative design variables and $\boldsymbol{t} = [t_1, t_2, \dots, t_m]^T$ are qualitative variables. Each qualitative design variable $t_j$ has $l_j$ design options (levels), i.e., $t_j \in \{1, 2, \dots, l_j\}$ for $j = 1, 2, \dots m$. For real physical models, there are quantitative variables $\boldsymbol{v}(t_j) = [v_1(t_j), v_2(t_j) \dots, v_n(t_j)] \in R^n$ underlying each qualitative variable that explains their influence on the response of interest, though these quantitative variables are usually unknown, not observable, or extremely high-dimensional. The key idea of LVGP is to learn a low-dimensional latent space to approximate the space of underlying quantitative variables $\boldsymbol{v}(t_j)$ via statistical inference. Although the dimensions of the latent variable vector $z \in R^k$ can be freely chosen, a two-dimensional (2D) latent vector, $k = 2$, is usually sufficient in most engineering designs [58][59], which is also adopted in this study. Thus, each level of a qualitative variable $t_j$ is represented by a 2D latent vector $\boldsymbol{z}(t_j) = [z_{j,1}(t_j), z_{j,2}(t_j)]^T$. The transformed design space becomes $\boldsymbol{h} = [\boldsymbol{x}^T, \boldsymbol{z}(\boldsymbol{t})^T] \in R^{(q+m \times 2)}$, where $\boldsymbol{z}(\boldsymbol{t}) = [z_{1,1}(t_1), z_{1,2}(t_1), z_{2,1}(t_2), z_{2,2}(t_2), \dots, z_{m,1}(t_m), z_{m,2}(t_m)]^T$.

Now, consider a single response GP model with a prior constant mean $\mu$ to describe the mean response at any given point in the design space $h$. A zero-mean Gaussian Process is used to capture the variance of the response, described by a covariance function $K(\boldsymbol{h}, \boldsymbol{h}')$. The covariance function $K(\boldsymbol{h}, \boldsymbol{h}') = \sigma^2 \cdot r(\boldsymbol{h}, \boldsymbol{h}')$ describes the relationship or the correlation of responses at any pairs of input points $h$ and $h'$, where $\sigma^2$ represents the prior variance of the GP model and $r(\boldsymbol{h}, \boldsymbol{h}')$ is the correlation function. LVGP extends the commonly used Gaussian correlation function to include latent variables,

$$r(\boldsymbol{h}, \boldsymbol{h}') = \exp\left(\sum_{i=1}^{q} \phi_i (x_i - x_i')^2 - \sum_{j=1}^{m} \left\| z_{j,1} - z_{j,1}' \right\|_2^2 + \left\| z_{j,2} - z_{j,2}' \right\|_2^2\right), \tag{1}$$

where $\phi_i$ is a scaling parameter that will be estimated for each quantitative variable $x_i$. The mapping from qualitative variables to 2D latent variables is scaled so that the scaling parameters of latent variables $\boldsymbol{z(t)}$ is a unity vector, which is set to be $\boldsymbol{1}$ as they will be estimated as hyperparameters. The rationale behind this correlation function is that points closed in the design space $\boldsymbol{h}$ should also exhibit similar output patterns. For a given design space with $n$ number of points, the parameters, $\mu, \sigma$, and $\phi$, along with the 2D mapped latent variables, $\boldsymbol{z(t)}$, are estimated through Maximum Likelihood Estimation (MLE), i.e., finding parameters to maximize the log-likelihood function,

$$l(\mu, \sigma, \boldsymbol{\phi}, \boldsymbol{z}) = -\frac{n}{2}\ln(2\pi\sigma^2) - \frac{1}{2}\ln|\mathbf{R}(\boldsymbol{z}, \boldsymbol{\phi})| - \frac{1}{2\sigma^2}(\boldsymbol{y} - \mu\mathbf{1})^T \mathbf{R}(\boldsymbol{z}, \boldsymbol{\phi})^{-1}(\boldsymbol{y} - \mu\mathbf{1}), \tag{2}$$

where $\mathbf{R}$ is the $n \times n$ correlation matrix with $\mathbf{R}_{ij} = r(h^i, h^j)$ for $i, j = 1, 2, 3, \ldots, n$, $\boldsymbol{1}$ is a vector of ones with dimensions of $n \times 1$, and $y = [y_1, y_2, \ldots, y_n]^T$ is the observed response vector. The MLE of the hyperparameters $\mu$ and $\sigma^2$ can be obtained analytically as:

$$\hat{\mu} = (\mathbf{1}^T \mathbf{R}^{-1} \mathbf{1})^{-1} \mathbf{1}^T \mathbf{R}^{-1} \boldsymbol{y} \tag{3}$$

$$\hat{\sigma}^2 = \frac{1}{n}(\boldsymbol{y} - \hat{\mu}\mathbf{1})^T \mathbf{R}^{-1}(\boldsymbol{y} - \hat{\mu}\mathbf{1}). \tag{4}$$

By substituting $\hat{\mu}$ and $\hat{\sigma}^2$ into Equation (2) and neglecting constants, the log-likelihood function becomes:

$$l(\boldsymbol{\phi}, z) \sim -n\ln(\hat{\sigma}^2) - \ln|\mathbf{R}(\boldsymbol{z}, \boldsymbol{\phi})|, \tag{5}$$

that depends on the scaling factor $\boldsymbol{\phi}$ and the latent variable $\boldsymbol{z}$. Once $\boldsymbol{\phi}$ and $\boldsymbol{z}$ are obtained by the MLE, the predictive mean of the LVGP at the arbitrary new point $\boldsymbol{x}^*$ and the variance of the error of this prediction is:

$$\hat{y}(\boldsymbol{x}^*) = \hat{\mu} + \boldsymbol{r}(\boldsymbol{x}^*)\mathbf{R}^{-1}(\boldsymbol{y} - \hat{\mu}\mathbf{1}) \tag{6}$$

$$\hat{s}^2(\boldsymbol{x}^*) = \hat{\sigma}^2(\boldsymbol{r}(\boldsymbol{x}^*) - \boldsymbol{r}(\boldsymbol{x}^*)\mathbf{R}^{-1}\boldsymbol{r}(\boldsymbol{x}^*)^T). \tag{7}$$

Where $\boldsymbol{r}(\boldsymbol{x}^*) = [r(\boldsymbol{x}^*, \boldsymbol{x}^{(1)}), \cdots, r(\boldsymbol{x}^*, \boldsymbol{x}^{(n)})]$ is a vector of pairwise correlation between $\boldsymbol{x}^*$ and every training sample $\boldsymbol{x}^i, i = 1, \dots, n$.

## 2.2 Formulation of MF-LVGP

MF-LVGP can be realized by embedding the fidelity levels as an additional qualitative variable for each input sample. As a result, the MF-LVGP learns the correlations between fidelity models, encodes them into a latent space, and builds a single smooth response surface that includes samples from all the fidelities.

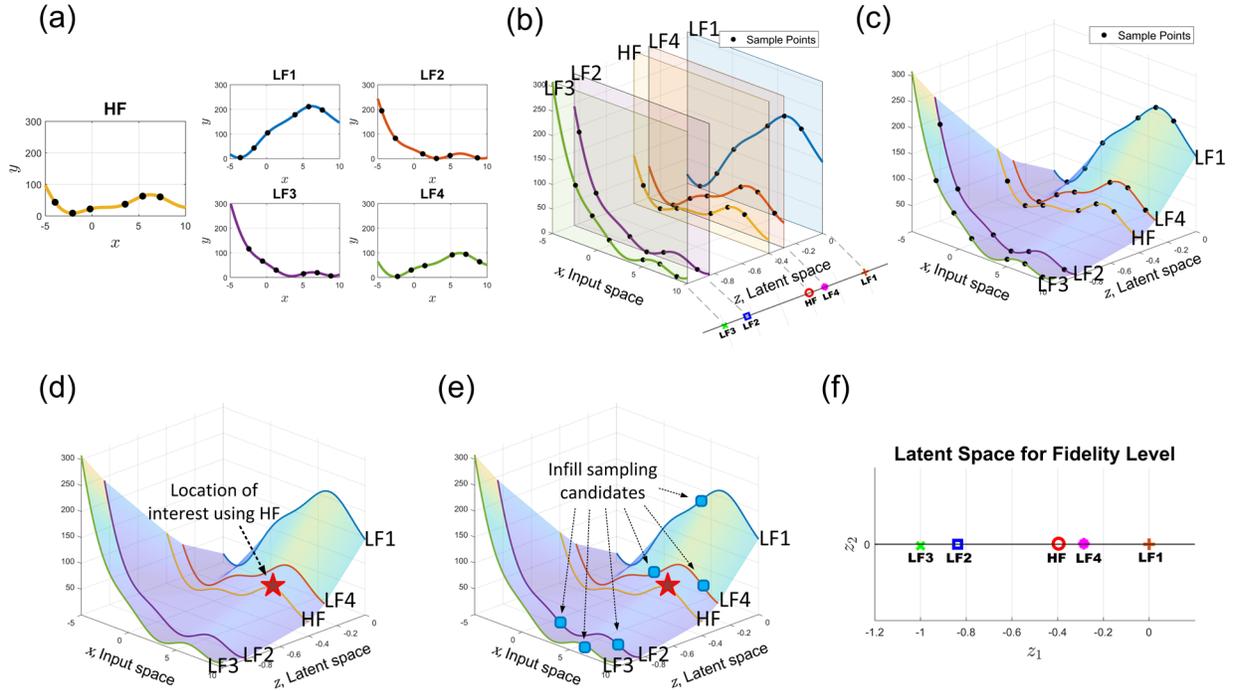

Figure 2: Illustration of MF-LVGP fitting and adaptive sampling: (a) Responses of each fidelity model. (b) The MF-LVGP learns the correlation between fidelity models by the additional latent dimension, where the distance between each model on the latent space reflects their correlations. (c) The MF-LVGP surrogates the MF models within a single surface by using latent space to quantify the correlation between fidelity. (d) The location of interest is the argmax of the acquisition function using HF surrogate model, identified in the first-stage optimization (e) The infill sampling candidates on LF models, where it can be distributed on all the fidelity models, and can be identified in the second-stage optimization. (f) The latent space of the fidelity level.

For MF modeling, besides the qualitative and quantitative variables introduced in the previous section, we use an extra variable $s$ to represent the fidelity level, i.e., $\boldsymbol{w} = [\boldsymbol{x}^T, \boldsymbol{t}^T, s]^T$. The key idea of MF-LVGP is to consider $s$ as a special qualitative variable and use the same inference technique to learn the latent representation of the fidelity level. Note that this MF formulation is first proposed by Eweis-Labolle *et al.* in [55] by using the LMGP, while in this work, we are using the LVGP, a more generalized representation of LMGP, to demonstrate the framework. However, by definition, we can still treat $s$ as a qualitative variable for the system and estimate its corresponding latent variable by replacing $\boldsymbol{t}$ by $\tilde{\boldsymbol{t}} = [\boldsymbol{t}^T, s]^T$ in the previous section. To compare our methods with the existing multi-fidelity methods that mostly can only handle quantitative parameters, in this work we only demonstrate examples where all the design/input parameters are quantitative. Therefore, for the rest of the paper, we focus on the scenarios with only quantitative variables and simplify the notation, i.e., $\boldsymbol{w} = [\boldsymbol{x}^T, s]^T$. However, we still note that our proposed method can be easily generalized to accommodate mixed variable problems by substituting $\boldsymbol{x}$ in the following equation as $\boldsymbol{w} = [\boldsymbol{x}^T, \boldsymbol{t}^T]^T$. Note that while assigning the fidelity level as a qualitative variable to the input samples, the modeler only needs to specify the highest fidelity model and the hierarchy of the LF model is not required. This is because the MF-LVGP will learn the correlation with the latent variables when fitting the surrogate model, and the fidelity level of LF models with respect to HF can be quantified and captured in the latent space.

In a MF system (number of fidelity models greater than three), we use a two-dimensional latent space to visualize the correlation between fidelity models. If two fidelity models are highly correlated, the distance between two fidelity models on the latent dimensions will be relatively small, and vice versa. An illustrative example of MF-LVGP is demonstrated in Figure 2. In this example, there are four one-dimensional LF models and one HF model, as visualized in Figure 2(a). When the MF-LVGP is trained, it learns the correlations between fidelity models on the additional latent space, as shown in Figure 2(f). Note that in this illustration we use a one-dimensional latent space for ease of visualizing the concept, while in practice we still suggest that users should use a two-dimensional latent spAace that fully captures the correlation between qualitative variables. Figure 2(b) shows how the relative correlation between fidelity models is quantified along the latent dimension. Models with highly correlated responses (e.g., HF and LF4, LF2 and LF3) are close to each other in the latent space, while weakly correlated models (e.g., HF and LF1, LF1 and LF3) are away from each other. While we do not assume any hierarchy among the LF models, the latent space can automatically infer and capture the possible hierarchy. For example, LF4 should be considered as an LF model with higher fidelity than LF1, since it is closer to and thus more correlated with HF. Finally, in Figure 2(c), we show that the MF-LVGP is treating the MF responses in a single, continuous surface based on the learned latent information. Figure 2(d) and Figure 2(e) will be introduced in the next section. This unique characteristic allows the MF-LVGP to identify the underlying fidelity level directly from the latent variables, and to address non-hierarchical MF problems if the latent space has more than one dimension. Built on the learned correlation, the novelty of this work lies in performing a new adaptive sampling strategy by explicitly utilizing the correlation represented by the latent embedding with the pre-posterior analysis illustrated in the following Section 2.3.

## 2.3 Pre-posterior Analysis

In this work, we implement the pre-posterior analysis of the LVGP model (also known as Kriging believers [57]) that uses a zeroth-order interpolation as the key component to bridging the LF and HF models in adaptive sampling. The zeroth-order interpolation assumes that the function value at the candidate point is known exactly and uses the posterior mean of the model as a surrogate for the function. This is a simple and fast way to estimate the utility of the point [60]. Specifically, for an arbitrary candidate sampling point, noted as $x_{next}$, we assume the predictive mean of its response $\hat{y}(x_{next})$ is an accurate approximation of its response and neglect the predictive uncertainty. We also assume the latent variables in Equation (1) remain constant while updating the pre-posterior model since they are estimated by the MLE and there is no closed-form representation to approximate the pre-posterior update.

The pre-posterior correlation matrix $\widehat{\mathbf{R}}_{new}$ is given in Equation (8) where $\mathbf{R}$ is the $n \times n$ correlation matrix and $r(\cdot,\cdot)$ is the correlation function in Equation (1). $\boldsymbol{x}$ and $\boldsymbol{y}$ are the initial training input and output vectors, respectively.

$$\widehat{\mathbf{R}}_{\boldsymbol{new}}(x_{next}) = \begin{bmatrix} \mathbf{R} & r(\boldsymbol{x}, x_{next})^T \\ r(\boldsymbol{x}, x_{next}) & r(x_{next}, x_{next}) \end{bmatrix}. \tag{8}$$

By assuming the $\widehat{\boldsymbol{y}}_{\boldsymbol{new}} = [\boldsymbol{y}, \hat{y}(x_{next})]^T$, the parameters in the LVGP can be updated as follows in Equation (9) and (10).

$$\hat{\mu}_{new}(x_{next}) = \left(1^T (\widehat{\mathbf{R}}_{\boldsymbol{new}}(x_{next}))^{-1} \mathbf{1}\right)^{-1} \mathbf{1}^T (\widehat{\mathbf{R}}_{new}(x_{next}))^{-1} \, \widehat{\boldsymbol{y}}_{\boldsymbol{new}} \tag{9}$$

$$\hat{\sigma}_{new}^2(x_{next}) = \frac{1}{n}(\widehat{\boldsymbol{y}}_{\boldsymbol{new}} - \hat{\mu}_{\boldsymbol{new}}(x_{next})\mathbf{1})^T \left(\widehat{\mathbf{R}}_{\boldsymbol{new}}(x_{next})\right)^{-1} (\widehat{\boldsymbol{y}}_{\boldsymbol{new}} - \hat{\mu}_{new}(x_{next})\mathbf{1}). \tag{10}$$

Thus, the pre-posterior predictive variance for an arbitrary location $\tilde{x}$ can be obtained as:

$$\hat{s}^2(\tilde{x}, x_{next}) = \hat{\sigma}_{new}^2(x_{next})(\boldsymbol{r}(\tilde{x}, x_{next}) - \boldsymbol{r}(\tilde{x}, x_{next})(\widehat{\mathbf{R}}_{\mathbf{new}}(x_{next}))^{-1} \boldsymbol{r}(\tilde{x}, x_{next})^{\mathrm{T}}). \tag{11}$$

We can also observe that once $x^*$ is selected, $\hat{s}^2(\tilde{x} | x_{next})$ becomes the function of $x_{next}$, i.e. we can represent how different $x_{next}$ influences the prediction uncertainty of $\tilde{x}$. In this work, we utilize them to perform a point-to-point quantification on how the infill sample of LF models will reduce the uncertainty/acquisition of the HF model. The details of this method are elaborated in the next section.

## 3. Multi-Fidelity Adaptive Sampling Framework (MuFASa)

In this section, we introduce a unified MF cost and benefit-aware adaptive sampling framework for both GF and BO. It takes advantage of the LVGP embedding to capture the correlation between fidelities and further uses the pre-posterior analysis to determine "*which fidelity*" and "*which location*" to sample the next data point.

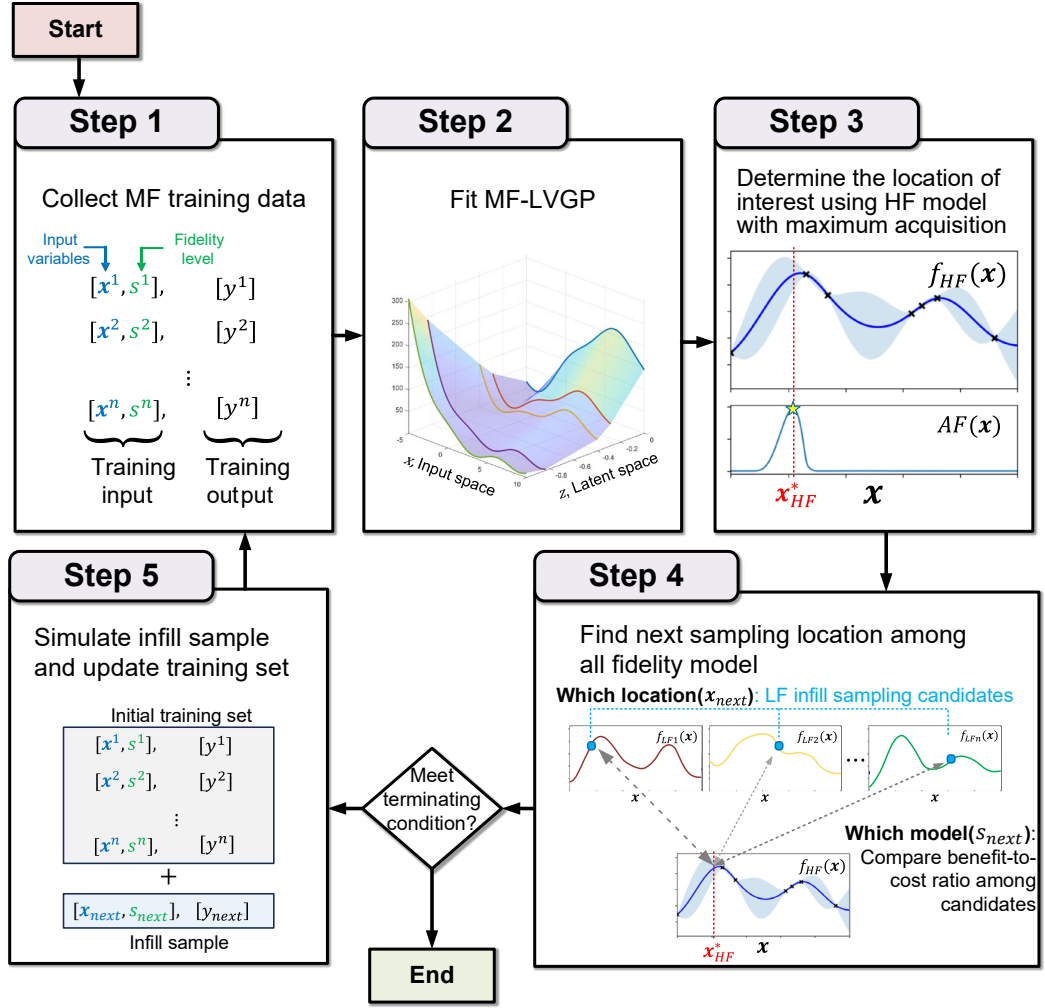

Figure 3: Flowchart of the MuFASa

### 3.1 Unified Framework for Global Fitting and Bayesian Optimization

The proposed framework follows the flowchart depicted in Figure 3. The method initiates with the design of experiment using low discrepancy sampling methods, such as the Sobol sequence [61] or Optimal Latin Hypercube Sampling [62]. After the initial sampling, the MF-LVGP is then trained by using the training dataset where the fidelity levels are assigned as the qualitative variable.

Next, our approach employs a two-stage optimization process to identify the optimal future sample (s). In the first-stage optimization, we identify the location of interest by using the HF surrogate model that yields the highest acquisition value. In the second-stage optimization, we explore all fidelity models to optimize the benefit-to-cost ratio that can be achieved by the future sample(s) to improve the location of interest obtained from the previous step. Once the locations of the future sample(s) are determined, we run the corresponding simulations or experiments based on the suggested sample location(s) with the corresponding fidelity model to add the next (or batch of) infill sample(s) into the training set. The MF-LVGP model is then refitted using the augmented training dataset until the terminating conditions are met, such as reaching the specified number of iterations or exhausting the computation budget.

The pseudo-code of the adaptive sampling with MF-LVGP is elaborated in Algorithm 1.

---

**Algorithm 1**: MuFASa

---

**Given:** Initial training set $\{w_{train} = [x_{train}, s_{train}], y_{train}\}$, HF model/source and LF model/sources $f_i(x), i = HF, LF1, \dots LF_n$, and the corresponding sampling cost $cost(i), i = HF, LF1, \dots LF_n$

**Goal:** Find the optimal solution of the HF model (BO); Improve the HF prediction (GF).

**Define:** AF, stopping criteria

**Step 0:** Train MF-LVGP with $\{w_{train}, y_{train}\}$

**While** *stopping criteria is not True* **do**

    1. First-stage optimization: Identify the location of interest $x^*_{HF}$ using HF model (Equation (12)):

$$x^*_{HF} = \arg\max AF(x_{HF})$$

    2. Second-stage optimization: Select an infill sample $w_{next}$ which includes the information of "which location" ($x_{next}$) and "which fidelity source" ($s_{next}$) with the best benefit-to-cost ratio (Equation (19)):

$$w_{next} = \arg\max \frac{\Delta \widehat{AF}(x^*_{HF}, w_{next})}{O(w_{next})}$$

    3. Simulate $f_i(x)$ at $w_{next}$ to obtain $y_{next}$, where $i$ indicates the selected source $s_{next}$.

    4. Update training set $\{w_{train}, y_{train}\} \leftarrow \{w_{train}, y_{train}\} \cup \{w_{next}, y_{next}\}$

    5. Re-train MF-LVGP with $\{w_{train}, y_{train}\}$

**End**

**Output:** Updated training set, the optimal solution of HF model (BO), or improved HF surrogate model (GF).

---

## 3.2 First-Stage Optimization: Determine the Location of Interest on HF Model

As shown in Figure 2(d), the objective of the first-stage optimization is to identify the location of interest $x^*_{HF}$ using HF surrogate model:

$$x_{HF}^* = \arg\max AF(x_{HF}),\tag{12}$$

where the acquisition function $AF(x)$ is a general representation that can be tailored according to the specific objective. For instance, for GF problems, the feasible AFs [22] include Maximum Mean Square Error (MMSE), Expected Improvement for GF (EIGF), Maximum Expected Predicted Error (MEPE), etc., and for BO, popular AFs [21] include Expected Improvement (EI), Probability of Improvement (PI), and so on.

### 3.2.1 Acquisition Function for Global Fitting

For GF problems in this work, we choose the MMSE as the default AF, as it indicates where on the HF model has the greatest uncertainty and requires the infill samples to provide information with top priority. The formulation of MMSE is shown as follows, where $\hat{s}^2(x)$ is the predictive uncertainty on $x_{HF}$ obtained by the MF-LVGP:

$$x_{HF}^* = \arg\max \hat{s}^2(x_{HF}).\tag{13}$$

When implementing the MMSE as the acquisition function, we may encounter two issues: First, during the iterations, we may repeatedly select the same $x_{HF}^*$ when its uncertainty remains the greatest. In such cases, repeatedly exploiting at that $x_{HF}^*$ may not be the most efficient strategy. In fact, by fixing on the same $x_{HF}^*$, we will lose the chance to explore other locations of the HF model with high prediction uncertainty, and to allocate LF samples across the entire design space that help the MF-LVGP improve the learned correlation between the fidelity models. Second, the $x_{HF}^*$ might locates near the boundary of the design space, which is less preferred in most engineering applications.

To address these issues, we employ a weighted random sampling method when identifying $x_{HF}^*$, named as "shuffle" in this study. Specifically, instead of greedily selecting $x_{HF}^*$ by maximizing $\hat{s}^2$ in Equation (13), we extract the modes of the MSE across the entire input space and take a weighted random sampling among the peaks $\hat{s}^2$ of different modes, with the respective peak values as the weights, to get the final $x_{HF}^*$. This shuffling mechanism forces the $x_{HF}^*$ to be selected from the modes, lower the chance of repeatedly selecting the same location as $x_{HF}^*$, and allocate the LF infill samples more evenly. To lower the computational cost and numerical tractability in high-dimensional problems, we use a multi-start gradient descent optimization to collect the unique local minimums as the modes instead of using Markov Chain Monte Carlo (MCMC) sampling method. We will demonstrate the ablation of the shuffling mechanism in the next section. Finally, we note that the weighted random selection of $x_{HF}^*$ among acquisition function modes is only used when the MMSE is chosen, as other functions like MEPE and EIGF already incorporate the balancing between exploration/exploitation.

### 3.2.2 Acquisition Function for Bayesian Optimization

For the BO problems in this work, we select the EI as our acquisition function. Like the MMSE we designed, the EI balances the exploration and exploitation while searching on the HF model, providing the location that has the highest potential to identify the optimum. EI of the HF model is formulated as [21]:

$$EI(x_{HF}) = (\hat{y}(x_{HF}) - y^*)\Phi\left(\frac{\hat{y}(x_{HF}) - y^*}{\hat{s}(x_{HF})}\right) + \hat{s}(x)\varphi\left(\frac{\hat{y}(x_{HF}) - y^*}{\hat{s}(x_{HF})}\right), \tag{14}$$

where $y^*$ is the current best solution within the training set $y_{train}$, $\Phi$ is the cumulative distribution function (CDF) of the quantity $z = \frac{\hat{y}(x_{HF}) - y^*}{\hat{s}(x_{HF})}$ and $\varphi$ is the probability density function (PDF), and $\hat{y}$ and $\hat{s}$ are the predicted mean response and predicted uncertainty from the LVGP, respectively. The $x_{HF}^*$ can be obtained via:

$$x_{HF}^* = \arg\max EI(x_{HF}). \tag{15}$$

While applying EI, one may encounter the similar issue as in the global fitting (GF) problem that the $x_{HF}$ remains the same during iterations. In this situation, we do not activate the shuffle mechanism because 1) the shuffle is to force exploration while the EI itself has the built-in exploration term and 2) in BO, if $x_{HF}^*$ is very close to the true optimum, we prefer to keep exploiting at $x_{HF}^*$.

## 3.3 Second-Stage Optimization: Determine the Most Cost-Efficient Infill Sample

The purpose of the second-stage optimization is to determine the infill sample $x_{next}$ that can improve the acquisition of $x_{HF}^*$ to the most with the best benefit-to-cost ratio. Here we provided two approaches to realize this.

**Approach 1**: Most Efficient Reduction in AF

We denote the acquisition of $x_{HF}^*$ as $AF(x_{HF}^*)$, and denote the pre-posterior acquisition, i.e. the approximate acquisition of $x_{HF}^*$ by assuming the infill sample $w_{next}$ is added to the training set and refit the MF-LVGP, as $\widehat{AF}(x_{HF}^*, w_{next})$. For example, in GF when the AF is MMSE, $\widehat{AF}(x_{HF}^*, w_{next})$ can be represented as:

$$\Delta\widehat{AF}(x_{HF}^*, w_{next}) := \Delta\widehat{MSE}(x_{HF}^*, w_{next}) = \hat{s}^2(x_{HF}^*) - \hat{s}^2(x_{HF}^*, w_{next}), \tag{16}$$

Where $\hat{s}(x_{HF}^*, w_{next})$ is the standard deviation of the pre-posterior distribution in Equation (12). In BO where the AF is EI, $\widehat{AF}(x_{HF}^*, w_{next})$ will become:

$$\Delta\widehat{AF}(x_{HF}^*, w_{next}) := \Delta\widehat{EI}(x_{HF}^*, w_{next}) = EI(x_{HF}^*) - \widehat{EI}(x_{HF}^*, w_{next}), \qquad (17)$$

$$\widehat{EI}(x_{HF}^*, w_{next}) = (\hat{y}(x_{HF}^*) - y^*)\Phi\left(\frac{\hat{y}(x_{HF}^*) - y^*}{\hat{s}(x_{HF}^*, w_{next})}\right) + \hat{s}(x, w_{next})\varphi\left(\frac{\hat{y}(x_{HF}^*) - y^*}{\hat{s}(x_{HF}^*, w_{next})}\right). \qquad (18)$$

Once the improvement of AF is determined, the second-stage optimization can be formulated as:

$$w_{next} = \arg\max \frac{\Delta\widehat{AF}(x_{HF}^*, w_{next})}{cost(w_{next})}, \qquad (19)$$

where $w_{next}$ is determined by searching through all the fidelity models, as shown in Figure 2(e). This formulation straightforwardly shows that the $w_{next}$ will provide the most improvement per unit cost.

However, in BO when the AF is EI, the Equation (18) becomes very complicated, leads to the vanishing gradient of the objective function, and as a result $\widehat{EI}$ loses the accuracy because the pre-posterior analysis only provides the approximation of the posterior MSE. Moreover, the complex objective function will dramatically decrease the applicability of optimizers in higher dimensional problems. Thus, an alternative method is proposed.

**Approach 2**: Most Efficient Reduction in $\hat{s}(x)$

Since the pre-posterior analysis only approximates the uncertainty of the pre-posterior model, in this approach, we directly replace the $\Delta\widehat{AF}$ with $\Delta\widehat{MSE}$ in Equation (19) as the unified objective function for this stage as:

$$w_{next} = \arg\max \frac{\Delta\widehat{MSE}(x_{HF}^*, w_{next})}{cost(w_{next})}, \qquad (20)$$

The mathematical meaning behind this approach is to say that once the $x_{HF}^*$ is selected by arbitrary AFs, the LF infill samples are providing information to lower the uncertainty at $x_{HF}^*$ before its HF infill sample is taken. This not only allows the optimizer to function well but can also prevent the BO from exploiting on the wrong optimum as the uncertainty around $x_{HF}^*$ is decreased.

Once $w_{next}$ is identified, we realize $w_{next}$ on the corresponding models, and include the $w_{next}$ and its output $y_{next}$ into the training set. We then refit the MF-LVGP and iterate through the same process until the stopping criteria is met. Note that the $w_{next}$ is not limited to represent one infill location, but it can also represent a batch of infill samples. It can be realized by 1) selecting several local optimums while optimizing Equation (19) or Equation (21) or 2) iteratively updating the posterior MF-LVGP and with predicted response of $\hat{y}_{next}$ and $w_{next}$ (see [60] for details about batch sampling with pre-posterior analysis). In this work, we only demonstrate one infill sample in each iteration.

### 3.4 Theoretical Analysis of MuFASa

Before diving into case studies, we provide a theoretical comparison between MuFASa and existing methods to unveil the effect of the key contribution of the proposed framwork, i.e., the pre-posterior analysis. The reason for including the pre-posterior analysis in the AF is to quantify how each LF sample candidate potentially benefits the HF surrogate. Specifically, most existing methods, such as MFCA[41], estimate the model uncertainty at a query point with Equation (7), which is only a function of the query input itself only, because each term in Equation (7) only involves the correlation between the query point and the existing dataset. As a result, when integrating it into an acquisition function, the acquisition function can only quantify how much a future sample can improve its own performance, but not the gain in other points or sources. In the context of the multi-fidelity problem, it indicates that the benefit/interplay between the candidate LF sample and the HF surrogate can not be captured in the acquisition function, which may mislead the adaptive sampling process. For example, if the LF sources are biased, i.e., have a different optimum from the HF sources, it may end up 1) taking samples on the biased LF sources because of the greater acquisition value, or 2) being misled by the biased LF sources and exploiting the incorrect region on the HF source, and converges to the incorrect optimum.

In contrast, using pre-posterior analysis in MuFASa, we essentially extend the model uncertainty estimation in Equation (7) to include both the query point and the future sample as the input in Equation (11). This is made possible by augmenting an extra term related to the future sample in both the column vector $\boldsymbol{r}$ and correlation matrix $\hat{\mathbf{R}}_{new}$. As a result, when using it in the acquisition function, it can capture the influence of a future sample on the gain at any point of interest. In the context of multi-fidelity problem, it means that we can quantitatively estimate the benefit of the LF sources on the HF points. Considering the aforemention scenario with biased LF sources, MuFASa will avoid selecting samples from the biased LF sources, because their contribution is penalized by their weak correlation with HF captured in the augmented terms in Equation (11). We will demonstrate the above theoretical analysis in the upcoming case studies, verifying that the utilization of correlation as the key information to navigate the sampling strategy can provide better sampling efficiency and correctness.

## 4. Case Study

In this section, we first present two 1D examples to illustrate how our methods work for GF and BO problems, respectively. We then apply our methods to two high-dimensional scenarios, showcasing the applicability of MuFASa in practical engineering problems.

We conduct ablation tests to demonstrate the effectiveness and contribution of each component in our framework. The details of the test cases for GF and BO are elaborated in Table 1 and Table 2 respectively. Note that the implementation of Approach 1 in Section 3.3 only appears in the 1D illustrative example but not in the high-dimensional test cases due to the complexity of the objective function. The terminating conditions for all the test cases are listed in Table 7.

Table 1: Descriptions of the test methods for GF

| Method | MF Data Fusion | Pre-posterior Analysis | Shuffle | Description |
|---|---|---|---|---|
| SFGP | [1]X | X | X | A single fidelity GP that only emulates the HF response and infill HF sample, navigated by single fidelity AFs (i.e., the first stage optimization) |
| MF-LVGP-HF | [2]O | X | X | An MF-LVGP is trained with data from all the fidelity sources, but only infills HF samples, navigated by the single fidelity AFs. |
| MuFASa-$\alpha$ | O | O | X | An MF-LVGP is trained with data from all the fidelity sources and infills on MF samples. The shuffle in the first-stage optimization is deactivated. |
| MuFASa-$\beta$ | O | O | O | An MF-LVGP is trained with data from all the fidelity sources and infills on MF samples, The shuffle in the first-stage optimization is activated. |
| CoKriging | Benchmark Method | | | A CoKriging is trained to emulate MF training samples, and the infill samples are determined by the SEVR proposed by [40]. See details in Algorithm A. |

Notes: [1]The feature is not demonstrated within the method, and [2]The feature is demonstrated within the method

Table 2: Descriptions of the test methods for BO

| Method | MF Data Fusion | Pre-posterior Analysis | AFs | Description |
|---|---|---|---|---|
| SFGP | [1]X | X | 1st: EI<br>2nd: --- | A single fidelity GP that only emulates the HF response and infill HF sample, navigated by single fidelity AFs (i.e., the first stage optimization) |
| MF-LVGP-HF | [2]O | X | 1st: EI<br>2nd: --- | An MF-LVGP is trained with data from all the fidelity sources, but only infills HF samples, navigated by the single fidelity AFs. |
| MuFASa-A | O | O | 1st: EI<br>2nd: $\Delta\widehat{AF}$ | An MF-LVGP is trained with data from all the fidelity sources and infills on MF samples. The AF in the second-stage optimization is the maximum decrease of the AF (Approach 1) |
| MuFASa-M | O | O | 1st: EI<br>2nd: $\Delta\widehat{MSE}$ | An MF-LVGP is trained with data from all the fidelity sources and infills on MF samples, The AF in the second-stage optimization is the maximum decrease of the MSE (Approach 2) |
| MFCA | Benchmark Method | | | An MF-LVGP is trained to emulate MF training samples, and the infill samples are determined by the multi-fidelity cost-aware AF proposed by [41]. See details in Algorithm B. |

Notes: [1]The feature is not demonstrated within the method, and [2]The feature is demonstrated within the method

## 4.1 Illustrative Example for Global Fitting

We implement the MuFASa to adaptively improve the global fitting performance on the Simple-1D

example modified from [55]. The Simple-1D has one HF model and three LF models, where the sampling costs on the LF models are identical (See other details in Table 3). Figure 4 shows the convergence history of MuFASa-$\beta$ in a single replicate, where we can see that the HF prediction in the first row gradually improves as the number of iterations increases. In this study, we employ the phrase 'rate of convergence' to describe qualitatively the speed at which the algorithm reaches a specific point, such as the potentially minimal RRMSE in GF or the optimal solution ($y^*$) in BO.

To begin with, at $n_{iter} = 0$, 2 HF samples and 15 LF samples (5 for each LF model) are generated to fit the initial surrogate. At $n_{iter} = 3$, the LF infill samples are taken on LF1 and LF2, leading to a significant increase in HF prediction. Note that the next $x^*_{HF}$ is not the exact maximum of the predictive uncertainty since the "shuffle" is activated to avoid being trapped in the same region of interest. At $n_{iter} = 6$, the MuFASa-$\beta$ keeps taking LF samples on LF1 and LF2, and one can observe that the uncertainty within the input space continues to decrease, and the predictive accuracy is improved significantly even without any infill on the HF. This is because the infill sample not only improves the individual LF surrogate model but also helps to better capture the correlation between fidelities in the latent space, thus improving the model performance on other fidelities, especially HF. This idea can also be verified at $n_{iter} = 9$, where the biased HF prediction at the region $x = [-2, -1]$ from $n_{iter} = 6$ is ameliorated by just infilling LF samples within that region instead of querying HF samples.

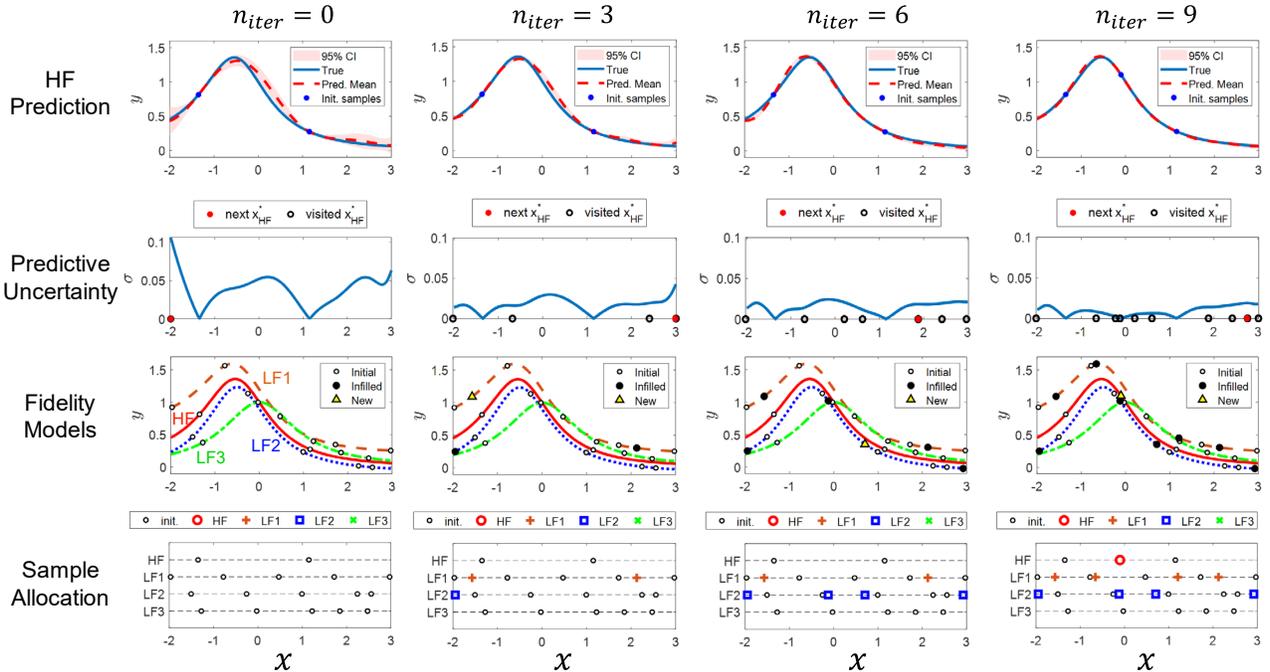

Figure 4: Iteration of MuFASa-$\beta$ in a single replicate for the Simple-1D problem.

We can also see that within these iterations, no infill sample has been taken on LF3, whereas more samples have been collected on LF2 than those on LF1. This showcases MuFASa's capability in correctly quantifying the benefit of future infill samples. Specifically, the latent representation and pre-posterior analysis automatically diminish the benefit when the infill candidate originates from an LF model less correlated with HF, indicating a smaller contribution in improving the HF prediction. This can be directly observed in the latent space of Figure 5(b), in which LF3 and HF are less correlated than that of LF1 and LF2 because the latent distance is larger. This hierarchy can be verified by the greater RRMSE of LF3 in Table 3.

Furthermore, we conducted 50 replicates of simulations on Simple-1D to assess the robustness of MuFASa. The resulting outcomes are then compared with those of other testing methods, as illustrated in Figure 5. In this case, MuFASa-$\alpha$ and MuFASa-$\beta$ yield similar performance in the rate of convergence and sample allocation, and both outperform other methods with more efficient convergence and much higher accuracy. Figure 5(a) and Figure 5(b) shows the individual/the median and the standard deviation of convergence history among the replicates. In Figure 5(a), before any samples are infilled, the RRMSE of MF-LVGP is already dramatically lower than SFGP, demonstrating the benefit of the latent representation of the MF system. In Figure 5(a) and Figure 5(d), CoKriging performs worse than all other methods in the first half of the infill sampling, including SFGP, primarily due to its hierarchical structure that leads to undesirable uncertainty accumulation along the fidelity levels while making HF prediction. Specifically, the sample size is not sufficient for CoKriging to sequentially fit the individual GPs for each fidelity model within its hierarchical framework (In this case, four GPs are trained in CoKriging). As the infill process is terminated and the HF information becomes abundant, the RRMSE of MF-LVGP-HF is worse than that of SFGP while MuFASa-α and MuFASa-β still maintain their superior performance. This outcome is attributed to the fact that when we solely infill HF samples in MF-LVGP-HF, the initial samples on LF models are inadequate for the MF-LVGP to accurately learn correlations between fidelities. In Figure 5(c), the latent distances not only represent the correlation between LF and HF models, but also determine the sampling frequency on each LF source. Here, given the uniform cost across LF models, the choice of infill sampling is solely driven by uncertainty reduction. Notably, the sampling frequency depicted in Figure 5(e) adheres to the sequence: $n_{LF2} > n_{LF1} > n_{LF3}$, which is the same as the sequence of distance to HF in the latent space.

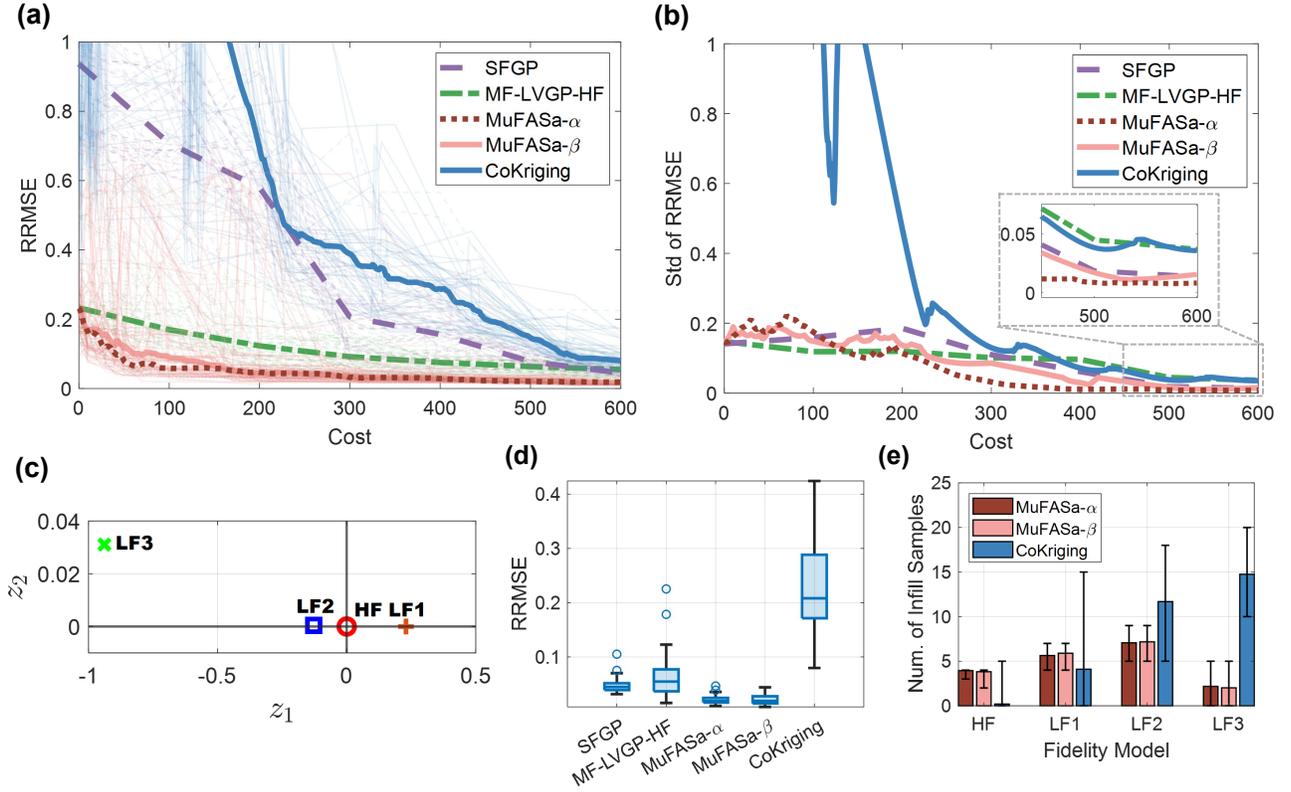

Figure 5: Results of 50 replicates of Simple-1D with the same computational budget for GF. (a) The convergence history of RRMSE vs. infill sampling cost (initial sample cost excluded). The thin lines represent different replicates, while the thick line is the median of the RRMSE at a given cost. (b) The standard deviation of the RRMSE vs infill sampling cost. (c) The latent space of the fidelity level from a randomly selected replicate. (d) The boxplot of RRMSE. (e) The infill sample allocation on fidelity sources of the MF test methods.

We note that the convergence histories of individual replicates do not decrease monotonically. This behavior arises from the variation of the estimated scaling factor through MLE, i.e., $\boldsymbol{\phi}$ in Equation (5). The inherent local variability within the training set significantly influences this scaling factor. Consequently, during the infill sampling stage, it is common to observe fluctuations in both SFGP and MF methods.

Concluding this example, the benefit of the MF latent representation is proved by comparing MF-LVGP and SFGP, and the advantage of adaptively selecting LF samples enabled by the pre-posterior analysis is highlighted while comparing MuFASa and the MF-LVGP-HF. We also show how our non-hierarchical framework outperforms the CoKriging with more efficient convergence.

## 4.2 Illustrative Example for Bayesian Optimization

In this section, we demonstrated the MuFASa in BO problem with the Sasena problem with three fidelities modified from [63]. This is also a non-hierarchical scenario where the two LF models have similar correlations to the HF model. Moreover, it challenges the robustness of the MF BO method because the optimum of the HF and LF models are different: The optimum of the LF models is located at $x = 2$ while that of the HF model is at $x = 8$. The details of the problem setting are listed in Table 4. Figure 6 shows the convergence process of MuFASa-M, elaborating how our method converges to the ground-truth optimum.

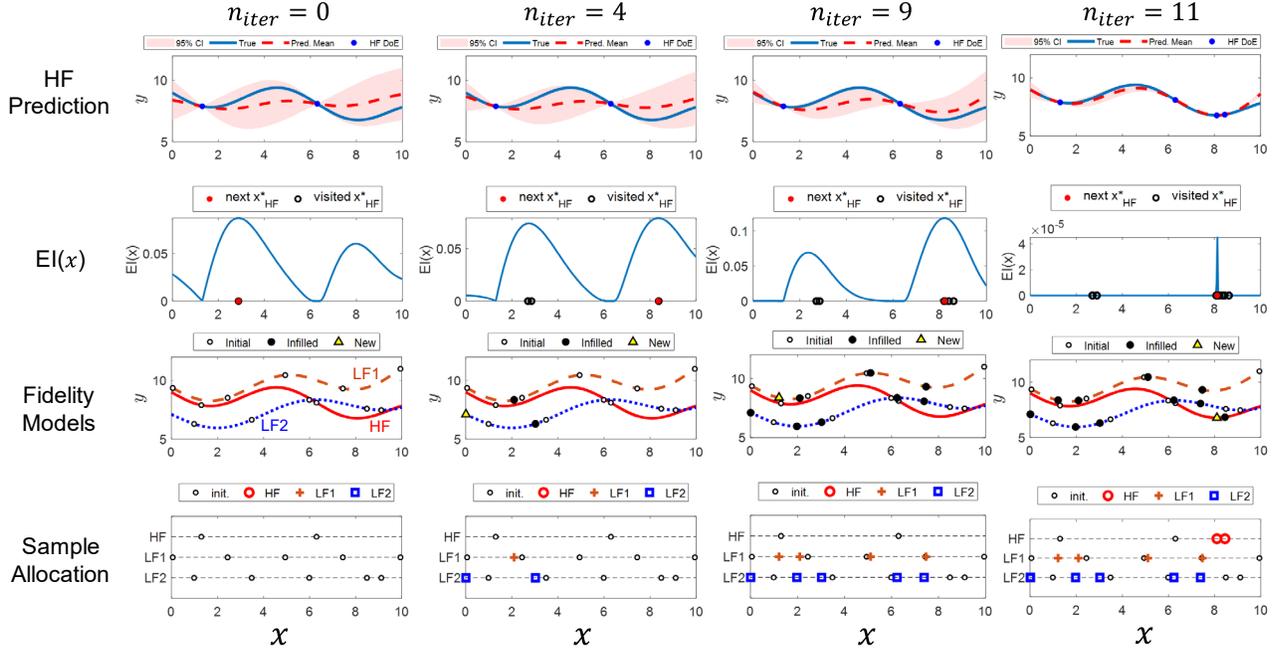

Figure 6: Iterations of MuFASa-M within a single replicate in case Sasena.

To begin with, in $n_{iter} = 0$, two HF samples and 10 LF samples (five for each LF model) are generated as the initial training set, and $x^*_{HF}$ is very close to the wrong optimum, i.e., the optimum on the LF instead of the HF. From $n_{iter} = 0$ to $n_{iter} = 4$, instead of exploiting around the wrong optimum using HF samples, the MuFASa-M takes 3 LF samples to reduce the uncertainty on $x^*_{HF}$. As a result, the $x^*_{HF}$ at $n_{iter} = 4$ moves toward the ground-truth optimum. From $n_{iter} = 4$ to $n_{iter} = 9$, the MuFASa-M continuously takes infill samples on LF models. Finally, at $n_{iter} = 10$ and $n_{iter} = 11$, two HF samples are infilled consecutively: The first HF sample lands around the ground-truth optimum, and the second HF sample successfully identifies the optimum to terminate the BO. Note that when the BO is terminated, the global model of the HF response is only accurate around the location of interest because the LF infill samples are selected specifically to reduce the uncertainty at the location of interest instead of the whole HF surrogate model. This highlights the role of the pre-posterior analysis in maximizing the sampling efficiency by enabling point-to-point uncertainty quantification to provide the most critical information for the location of interest.

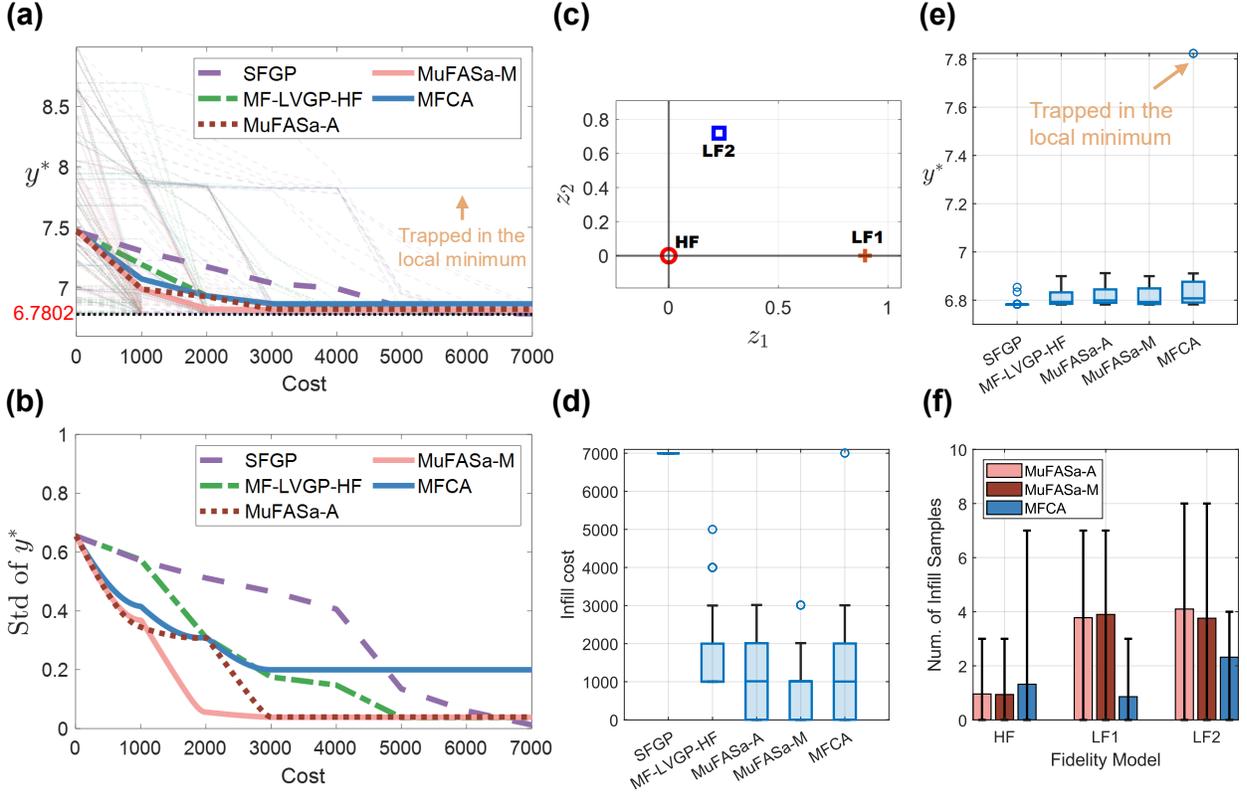

Figure 7: Results of 50 replicates of Sasena for BO. (a) The convergence history of HF optimal solution vs. infill sampling cost. The thin line represents the HF optimal solutions vs. cost for each iteration, and the thick line is the mean of the optimal solution vs. cost among all the replicates. The cost of initial samples is not included. (b) The standard deviation of optimal solution vs. cost among all the replicates. (c) The latent space of the fidelity level from a randomly selected replicate. (d) The boxplot of the infill cost. (e) The box plot of the optimal solution. (f) The infill sample allocation on fidelity sources.

Similarly, we performed 50 replicates to compare the robustness and efficiency of MuFASa with other methods. From Figure 7(a), to Figure 7(e), we can see that MuFASa-M outperforms all the other methods, converging to the correct optimum with the least sampling cost and on average. Also, we can see that the MF-LVGP-HF converges faster than SFGP, showing that the HF surrogate benefits from the latent representation of MF. The average rate of convergence of MuFASa-A is similar to that of MFCA, while it outperforms MFCA in terms of robustness, i.e., the latter has a chance to converge to the wrong optimum. This is because the AF of the MFCA on HF surrogate is the improvement (See Algorithm B) that does not consider the uncertainty at the location of interest and allows the BO to exploit the wrong optimum. On the other hand, our method constantly utilizes the information from the LF infill samples to reduce the uncertainty of the selected location of interest, maximizing the probability of exploiting the correct optimum. This can also be observed from Figure 7(f) that the MFCA takes fewer LF samples before exploitation while the MuFASa prefers to take more LF samples and requires fewer HF samples for exploration.

In this non-hierarchical scenario where LF1 and LF2 have similar correlations to the HF model, we can see in Figure 7(c), that the latent distance of LF1 and LF2 are nearly the same. Because the LF1 and LF2 are considered to have equal contributions to the HF, the sampling frequency from the LF source shown in Figure 7(f) is nearly identical between LF1 and LF2 for both MuFASa-A and MuFASa-M.

Concludingly, this illustrative example demonstrates how the MuFASa benefits from the pre-posterior analysis in achieving efficient and robust sampling strategies to navigate the BO. We also suggest using MuFASa-M rather than MuFASa-A for future cases in terms of the feasibility of the optimizer and the performance.

## 4.3 High-dimensional cases

In this section, we applied MuFASa in GF and BO on two high-dimensional cases: an 8D Borehole function and a 10D Wing Weight model, both modified from [41], with the detailed problem settings outlined in Table 5 and Table 6. We performed 20 replicates for both cases and compared with the test methods listed in Table 5 and Table 6. Note that in BO, we only applied MuFASa-M since MuFASa-A is not feasible in high-dimensional problems.

### 4.3.1 Global Fitting in High-Dimensional Cases

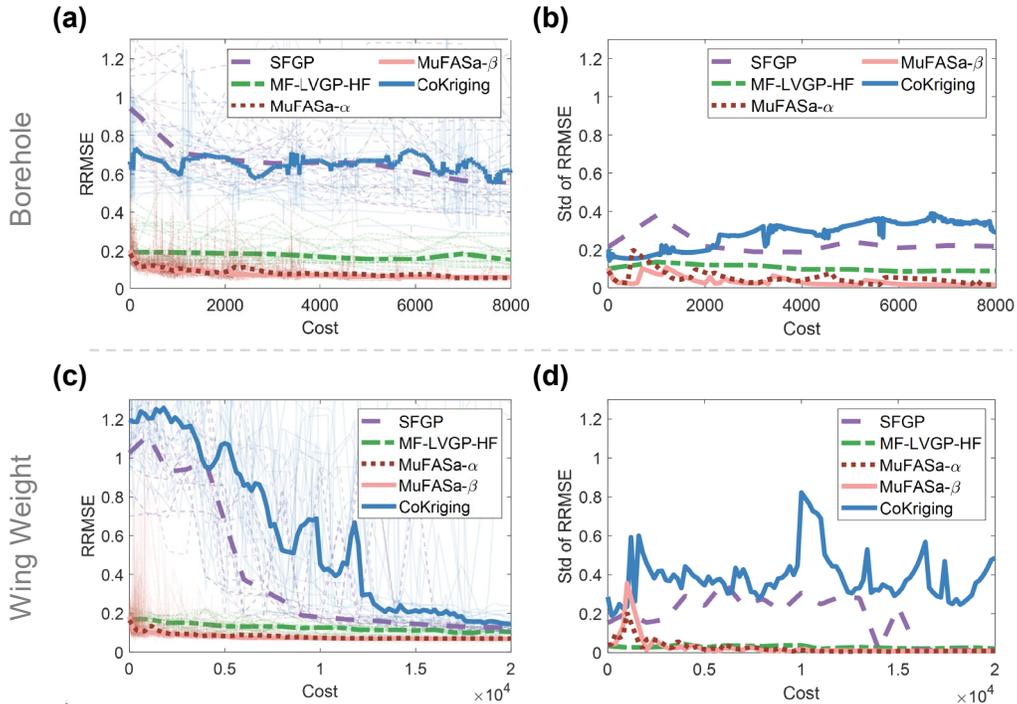

Figure 8: (a) Convergence history of 20 replicates of Borehole in GF. The thin line is the RRMSE vs. cost for each iteration, and the thick lines are the median of the RRMSE at given cost. (b) The standard deviation of RRMSE vs. cost in Borehole (c) Convergence history of 20 replicates of Wing Weight in GF. The thin line is the RRMSE vs. cost for each iteration, and the thick lines are the median of the RRMSE at given cost. (d) The standard deviation of RRMSE vs. cost in Wing Weight.

20 replicates of Borehole and Wing Weight are simulated for GF, and the results are shown in Figure 8 and Figure 9. From Figure 8 (a) to Figure 8(d), both MuFASa-$\alpha$ and MuFASa-$\beta$ converge faster than other methods. In Figure 8, for both scenarios, it's evident that the RRMSE reduces rapidly during the initial phase (cost < 2000), taking many LF samples in the early stages when the HF uncertainty is still large. The RRMSE quickly converges to a low level and remains stable even when more HF samples are infilled in the later stage (cost > 2000). We believe this phenomenon occurs because the surrogate model exhibits a certain level of accuracy to the extent that the newly introduced HF samples have minimal impact on the original distribution. Another explanation is rooted in the highly nonlinear nature of the response surface. As the additional HF samples primarily affect the local response, the RRMSE, which reflects the aggregate accuracy of the global model, might not effectively capture the localized enhancement in surrogate performance. What's more, the HF prediction might be inherently biased since it is conditioned on biased LF sources that undermine the accuracy of the prediction. Nevertheless, we still demonstrate the superiority of MuFASa in more efficient and accurate convergence compared to the benchmark methods.

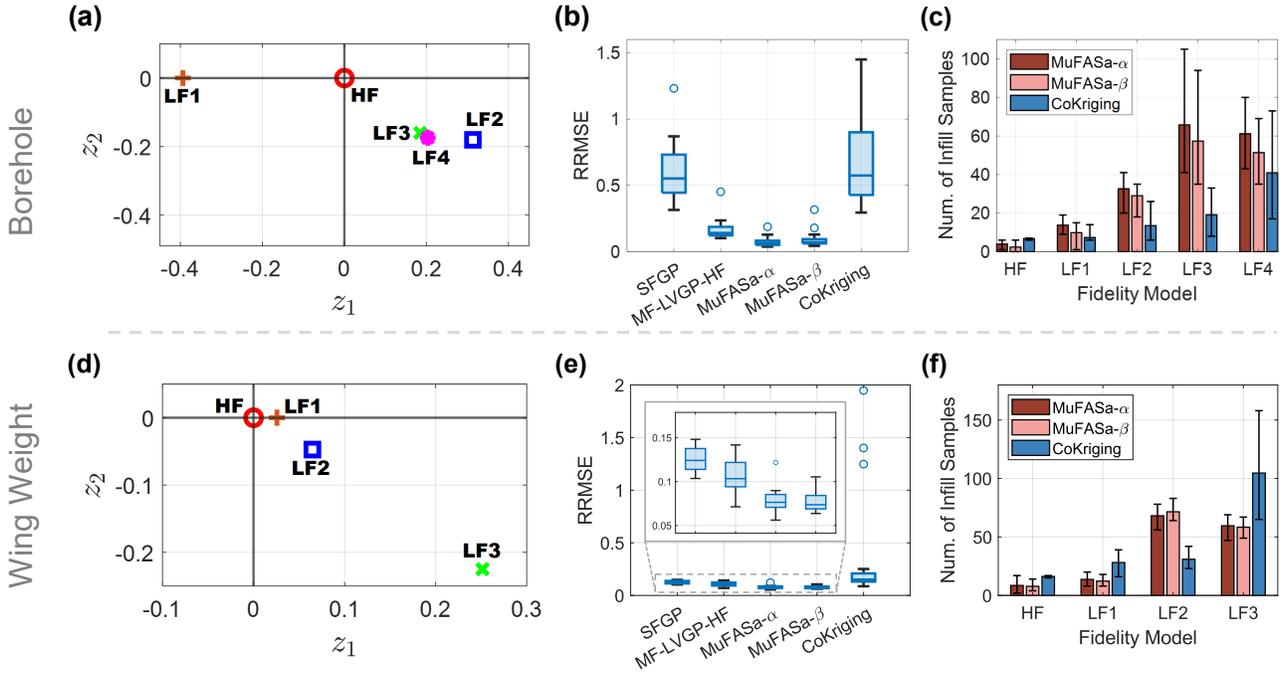

Figure 9: Results of 20 replicates of Borehole and Wing Weight model in GF. (a) The latent space of the fidelity level of Borehole from a randomly selected replicate. (b) The boxplot of RRMSE in Borehole. (c) The infill sample allocation on fidelity sources of the MF test methods in Borehole. (d) The latent space of the fidelity level of Wing Weight from a randomly selected replicate. (e) The boxplot of RRMSE in Wing Weight with the zoom region inset. (f) The infill sample allocation on fidelity sources of the MF test methods under in Wing Weight.

We observe that CoKriging fails to allocate sampling resources based on the correlation of the model but the hierarchy of the fidelity models. Since it must follow the nested data structure (See Algorithm A), the number of infill samples on the lower fidelity model will always be greater than the higher ones. On the other hand, in MuFASa-$\alpha$ and MuFASa-$\beta$, the sampling frequency on LF sources closely relates to the learned correlation between the LF and HF models. In the Borehole problem where each LF source has an equal sampling cost, LF3 and LF4 are the most visited sources since they have the shortest latent distances in Figure 9(a), i.e., have a higher correlation with the HF model. Moreover, in the Wing Weight problem where the sampling cost corresponds to the hierarchy, i.e., the sampling cost from the higher fidelity source is more expensive, the sampling frequency on LF sources reflects the compromised result between the correlation and the cost.

### 4.3.2 Bayesian Optimization in High-Dimensional Cases

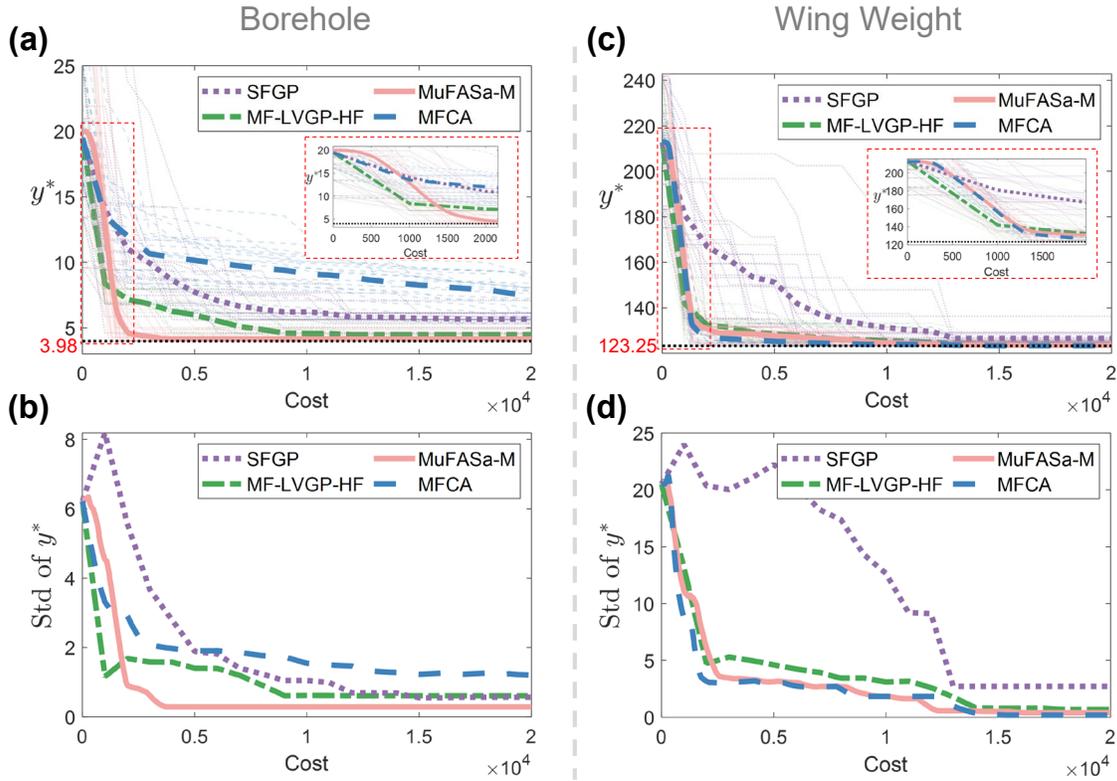

Figure 10: Convergence history of 20 replicates of Borehole and Wing Weight in BO. The black dash line indicates the ground truth optimal value of the HF model. The inset highlights the convergence history at the early stage of the BO. (a) Convergence history of 20 replicates of Borehole in BO. The thin line is the $y^*$ vs. cost for each iteration, and the thick lines are the mean of the $y^*$ at given cost. (b) The standard deviation of $y^*$ vs. cost of Borehole (c) Convergence history of 20 replicates of Wing Weight in GF. The thin line is the $y^*$ vs. cost for each iteration, and the thick lines are the mean of $y^*$ at given cost. (d) The standard deviation of $y^*$ vs. cost in Wing Weight.

We performed 20 replicates of BO for the Borehole and Wing Weight problems, with the results shown in Figure 10 and Figure 11. Overall, we can see that the MuFASa-M outperforms other competing methods in Borehole, which can find a better solution closest to the true optimum under a limited budget and iterations with less sampling cost. It exhibits comparable performance with MFCA in the Wing Weight scenario.

Upon closer examination of Figure 10, we can see that instead of exploiting the HF source immediately, the mean curve of MuFASa-M is flat at the beginning of the BO where $y^*$ remains constant. This is the stage where LF samples are infilled for MuFASa-M to identify the $x^*_{HF}$ around the ground-truth optimum. As a result, the MuFASa-M can approach the ground-truth optimal with its first few HF infill samples. In contrast, the MFCA failed to identify the correct optimum because its AF ignores the uncertainty of HF in Borehole. Although MFCA can exploits the HF model slightly faster and more accurate than MuFASa in some scenarios, e.g. in Wing Weight, see Figure 10(b) and Figure 11(e) , it puts the BO at the risk of wrong convergence once the LF model is highly biased.

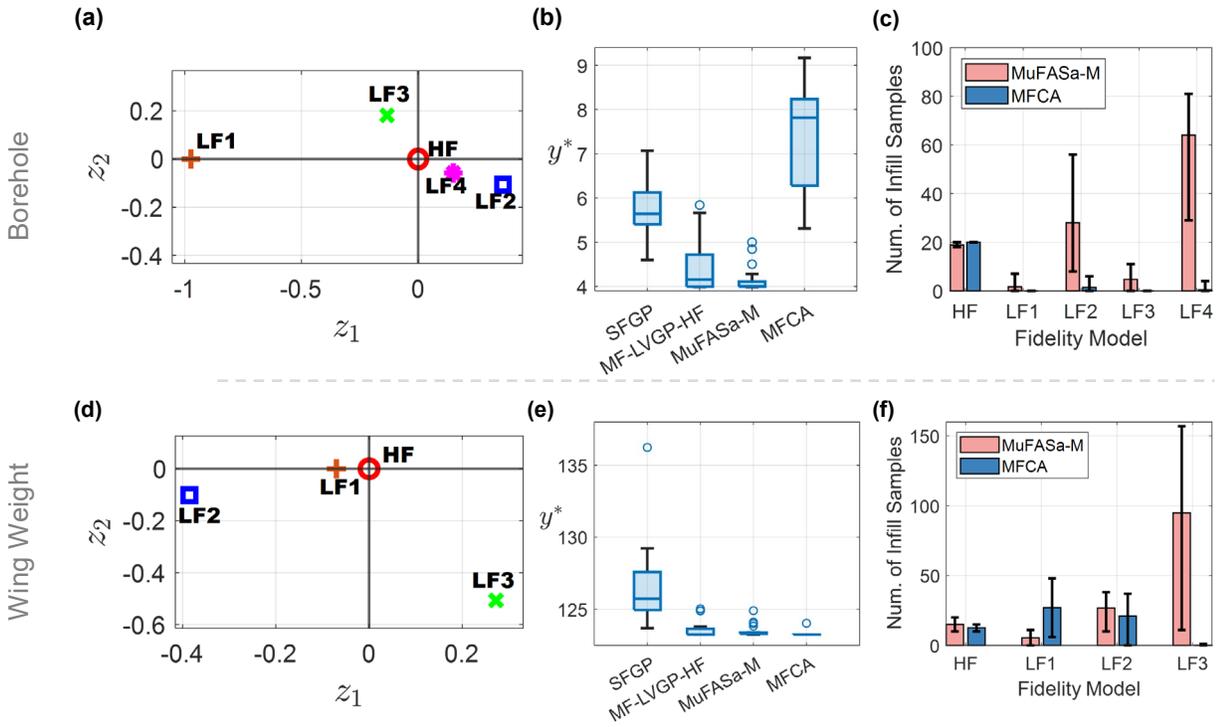

Figure 11: Results of 20 replicates of Borehole and Wing Weight model in BO. All the quantities are saved when the terminating conditions are met. (a) The latent space of the fidelity level of Borehole from a randomly selected replicate. (b) The boxplot of optimal solutions in Borehole. (c) The infill sample allocation on fidelity sources of the MF test methods in Borehole. (d) The latent space of the fidelity level of Wing Weight from a randomly selected replicate. (e) The boxplot of optimal solution in Wing Weight. (f) The infill sample allocation on fidelity sources of the MF test methods in Wing Weight.

Comparing the latent space of BO (Figure 11(a) and Figure 11(b)) and those of GF (Figure 9(a) and Figure 9(b)), it is apparent that the relative distances between the fidelity model markers are different, while the hierarchy learned by the MF-LVGP remains consistent. This observation indicates that the latent variables exhibit variations in response to changes in the training set (the union of the initial samples and the infill samples). Given that GF and BO employ distinct resource allocation strategies, it is plausible that the captured correlations also differ. By investigating the resource allocation behavior from Figure 11(c) and Figure 11(f), we can conclude that MuFASa-M tends to take more samples from cheaper LF sources. In contrast, more samples are taken on the more expensive but highly correlated LF sources for the MFCA. One can even observe that no samples have been taken from LF3 for BO with the MFCA. This may be because the rationale behind the AF of the two methods is different: The MFCA prefers to exploit the HF model and explore the individual LF models with higher expected improvement, while the MuFASa-M is driven by uncertainty reduction on HF prediction to guarantee successful exploitation. Thus, we can conclude that our method exhibits a comparable or better rate of convergence than the state-of-the-art MF methods but outperforms them in terms of robustness.

## 5. Conclusions

The major contribution of our work is to extend pre-posterior analysis and LVGP to develop a more efficient, flexible, and robust adaptive sampling framework for both GF and BO with non-hierarchical MF. To accommodate non-hierarchical scenarios where the rank of fidelity levels is unknown in prior, we introduce MF-LVGP, a method that statistically infers an interpretable latent representation to capture correlations among an arbitrary number of LF models. We further integrate pre-posterior analysis using zeroth-order interpolation with MF-LVGP to create the Multi-Fidelity Adaptive Sampling (MuFASa) framework that enables a sampling strategy considering both benefit and cost, explicitly leveraging the acquired correlations to quantify the future benefits of the candidate infill samples. Without the need to construct a new AF, our framework can accommodate any existing single-fidelity AF based on the intended use and offers the flexibility of switching the objective between GF and BO. The combination of LVGP and the pre-posterior analysis provides a new perspective on how interpretable correlations between fidelities can navigate the adaptive sampling strategies. Further, by comparing the convergence process on replicate tests, we demonstrate the superior convergence efficiency and robustness of MuFASa, while providing insights into the contributions of each pivotal component. Specifically, for GF, this method outperforms the CoKriging in the rate of convergence, exhibiting better predictive accuracy with less infill sampling cost; for BO, it outperforms the MFCA in terms of more efficient and correct convergence to the ground truth optimal. In both scenarios, we demonstrate how our method quantifies the benefit of the candidate infill samples, allocates appropriate sampling resources to each source, and makes decisions that factor in the sampling cost. On the application end, we expect that this method can contribute to a wide variety of practical engineering design problems, such as metamaterial system design where several fidelity models coexist [64], manufacturing systems where expensive experimental data are incorporated with simulation models [65], and polymer design where the simulations are obtained via different computational methods [66].

Nevertheless, some limitations still exist for the proposed method. As a common challenge for GP-based methods, its computational complexity hinders its scalability to big data. Also, as we only use stationary latent variables as an aggregate representation of the correlation, i.e., assuming that the correlation between fidelity models remains the same among the whole design space, it may not be able to capture nonstationary correlations. We assume the latent variables remain fixed in approximating the posterior distribution, which may lead to inaccurate pre-posterior analysis when there are many qualitative variables beyond the one that represents model fidelity. Further, in this work, we only demonstrate the case where the HF and LF models share the same input variables and input spaces, i.e., the upper and lower bound of the input variables are the same. This method is not applicable when the input space of LF models is just a subregion of that of the HF model.

These limitations also lead to our future endeavors to push the limit of this method. To tackle the large sample size and high-dimensionality issues, we can employ scalable techniques for GP such as neural processes and variational inference [67], manifold learning techniques like Uniform Manifold Approximation and Projection (UMAP) [68] and t-distribution Stochastic Neighbor Embedding (t-SNE) [69], and reduced order model (ROM) techniques [70] including input mapping [71] and Variational Autoencoders (VAE) [72], within the proposed framework. By inferring or constructing a unified representation space/latent space for all the sources, we expect that these approaches can alleviate the computational burden associated with the large dataset. It should also allow us to effectively reduce the dimensionality of inputs to the LVGP model, and enhance its feasibility for addressing higher-dimensional engineering problems and scenarios where the input variables across different fidelity models are different. Further, instead of learning a set of constant latent variables, we may construct the latent variables as a function of input variables in the future, which can capture more complex nonstationary correlations between different models or sources of data to improve the generality of this method. However, note that building this mapping function, e.g., by integrating with neural networks, will be at the cost of more samples and its cost-benefit will need to be investigated and compared with the current approach with the stationary assumption. Moreover, rather than solely opting for a single infill sample per iteration, the incorporation of more efficient strategies such as greedy sampling techniques [43], and dynamic programming-based approaches [73] can be implemented. By doing so, we can maximize the utilization of our budget by planning several steps ahead. These methodologies can be seamlessly integrated with the pre-posterior analysis, allowing us to approximate the benefits of future steps.

## Acknowledgements

We appreciate the grant support from the National Science Foundation Future Manufacturing Research Program (NSF 2037026) and the Engineering Research Center for HAMMER (NSF 2133630).

We also appreciate Professor Daniel Apley for the instruction on technical writing. Finally, we appreciate Zahra Zanjani Foumani, Amin Yousefpour, and Professor Ramin Bostanabad from University of California, Irvine for providing LMGP and MFCA codes.

# Appendix

## A.1 Benchmark Problems

Table 3 to Table 6 are the test cases demonstrated in this paper. The RRMSE (Relative Root Mean Squared Error) in each table is a matrix to quantify the difference between each LF source and the corresponding HF model, formulated as follows:

$$RRMSE = \sqrt{\frac{(\mathbf{y_l} - \mathbf{y_h})^T (\mathbf{y_l} - \mathbf{y_h})}{10{,}000 \times var(\mathbf{y_h})}},$$

Table 3: Formulation, sampling cost, and initial sample size of Simple-1D, modified from [55].

| Fidelity model | Formulation | Upper/lower bound | RRMSE | Cost | Init. Sample |
|---|---|---|---|---|---|
| HF | $y_{HF}(x) = \dfrac{1}{0.1x^3 + x^2 + x + 1}$ | | --- | 100 | 2 |
| LF1 | $y_{LF1}(x) = \dfrac{1}{0.2x^3 + x^2 + x + 1} + 0.2$ | $-2 \le x \le 3$ | 0.6054 | 10 | 5 |
| LF2 | $y_{LF2}(x) = \dfrac{1}{x^2 + x + 2} - 0.1$ | | 0.3218 | 10 | 5 |
| LF3 | $y_{LF3}(x) = \dfrac{1}{x^2 + 1}$ | | 0.7256 | 10 | 5 |

Table 4: Formulation, sampling cost, and initial sample size of Sasena, modified from [74] and [63]. The ground truth optimal value of HF is 6.7802.

| Fidelity model | Formulation | Upper/lower bound | RRMSE | Cost | Init. Sample |
|---|---|---|---|---|---|
| HF | $y_{HF}(x) = -\sin x - \exp\left(\dfrac{x}{10}\right) + 10$ | | --- | 1000 | 2 |
| LF1 | $y_{LF1}(x) = -\sin(0.95x) - \exp\left(\dfrac{x}{50}\right) + 0.03(x-2)^2 + 10.3$ | $0 \le x \le 10$ | 2.0544 | 1 | 5 |
| LF2 | $y_{LF2}(x) = -\sin(0.8x) - \exp\left(\dfrac{x}{50}\right) + 0.03(x-2)^2 + 8$ | | 1.8060 | 1 | 5 |

Table 5: Formulation, sampling cost, and initial sample size of Borehole [41]. The ground truth optimal value of HF is 3.98. Note that the setting in sampling cost and initial sample size are different in GF and BO. To obtain comparable results, we use the initial samples of {50,50,50,50,20} for {HF, LF1, LF2, LF3, LF4} for CoKriging.

| Fidelity model | Formulation | Upper/lower bound | RRMSE | Cost | | Init. DoE | |
|---|---|---|---|---|---|---|---|
| | | | | GF | BO | GF | BO |
| HF | $y_{HF} = \dfrac{2\pi T_u(H_u - H_l)}{ln\left(\frac{r}{rw}\right)(1 + \frac{2LT_u}{ln\left(\frac{r}{rw}\right)r_w^2 k_w} + \frac{T_u}{T_l})}$ | $100 \leq T_u \leq 1000$ | --- | 1000 | 1000 | 4 | 5 |
| LF1 | $y_{LF1} = \dfrac{2\pi T_u(H_u - 0.8H_l)}{ln\left(\frac{r}{rw}\right)(1 + \frac{1LT_u}{ln\left(\frac{r}{rw}\right)r_w^2 k_w} + \frac{T_u}{T_l})}$ | $990 \leq H_u \leq 1110$ $700 \leq H_l \leq 820$ | 3.6649 | 10 | 100 | 10 | 5 |
| LF2 | $y_{LF2} = \dfrac{2\pi T_u(H_u - H_l)}{ln\left(\frac{r}{rw}\right)(1 + \frac{8LT_u}{ln\left(\frac{r}{rw}\right)r_w^2 k_w} + 0.75\frac{T_u}{T_l})}$ | $100 \leq r \leq 10000$ $0.05 \leq r_w \leq 0.15$ | 1.3679 | 10 | 10 | 10 | 25 |
| LF3 | $y_{LF3} = \dfrac{2\pi T_u(1.09H_u - H_l)}{ln\left(\frac{4r}{rw}\right)(1 + \frac{3LT_u}{ln\left(\frac{r}{rw}\right)r_w^2 k_w} + \frac{T_u}{T_l})}$ | $10 \leq T_l \leq 500$ $1000 \leq L \leq 2000$ | 0.4135 | 10 | 100 | 10 | 5 |
| LF4 | $y_{LF4} = \dfrac{2\pi T_u(1.05H_u - H_l)}{ln\left(\frac{2r}{rw}\right)(1 + \frac{3LT_u}{ln\left(\frac{r}{rw}\right)r_w^2 k_w} + \frac{T_u}{T_l})}$ | $6000 \leq K_w \leq 12000$ | 0.4828 | 10 | 10 | 10 | 25 |

Table 6: Formulation, sampling cost, and initial sample size of Wing Weight model [55]. The ground truth optimal value of HF is 123.25. The settings are identical in GF and BO.

| Fidelity model | Formulation | Upper/lower bound | RRMSE | Cost | Init. Sample |
|---|---|---|---|---|---|
| HF | $0.036 s_w^{0.758} w_{fw}^{0.0035} \left(\frac{A}{cos^2(\Lambda)}\right)^{0.6} q^{0.006} \lambda^{0.04} \left(\frac{100 t_c}{cos(\Lambda)}\right)^{-0.3}$ $\times (N_z W_{dg})^{0.49} + s_w w_p$ | | --- | 1000 | 5 |
| LF1 | $0.036 s_w^{0.758} w_{fw}^{0.0035} \left(\frac{A}{cos^2(\Lambda)}\right)^{0.6} q^{0.006} \lambda^{0.04} \left(\frac{100 t_c}{cos(\Lambda)}\right)^{-0.3}$ $\times (N_z W_{dg})^{0.49} + w_p$ | $150 \leq s_w \leq 200$ $220 \leq w_{fw} \leq 300$ $6 \leq A \leq 10$ $-10 \leq \Lambda \leq 10$ $16 \leq q \leq 45$ $0.5 \leq \lambda \leq 1$ $0.08 \leq t_c \leq 0.18$ $2.5 \leq N_z \leq 6$ $1700 \leq W_{dg} \leq 2500$ $0.025 \leq w_p \leq 0.08$ | 0.1990 | 100 | 5 |
| LF2 | $0.036 s_w^{0.8} w_{fw}^{0.0035} \left(\frac{A}{cos^2(\Lambda)}\right)^{0.6} q^{0.006} \lambda^{0.04} \left(\frac{100 t_c}{cos(\Lambda)}\right)^{-0.3}$ $\times (N_z W_{dg})^{0.49} + w_p$ | | 1.1424 | 10 | 10 |
| LF3 | $0.036 s_w^{0.9} w_{fw}^{0.0035} \left(\frac{A}{cos^2(\Lambda)}\right)^{0.6} q^{0.006} \lambda^{0.04} \left(\frac{100 t_c}{cos(\Lambda)}\right)^{-0.3}$ $\times (N_z W_{dg})^{0.49}$ | | 5.7469 | 1 | 50 |

The terminating criterion for each case is listed in the following table:

Table 7: Terminating conditions for case study.

| Case | Problem Type | Terminating Condition |
|---|---|---|
| Simple-1D | GF | • $n_{iter} \geq 20$ or $cost \geq 600$ |
| Sasena | BO | • Error between optimal and the ground truth $\leq 2\%$ , or • $cost \geq 7000$ |
| Borehole | GF | • $n_{iter} \geq 150$ or $cost \geq 8000$ |
| | BO | • $n_{iter} \geq 200$ or $cost \geq 20,000$ |
| Wing Weight | GF | • $n_{iter} \geq 150$ or $cost \geq 20,000$ |
| | BO | • $n_{iter} \geq 200$ or $cost \geq 20,000$ |

## A.2 GF Benchmark Method: Adaptive sampling for CoKriging

In this work, we adapted the infill sampling framework for CoKriging as the benchmark method. The details of the MF data fusion of CoKriging the adaptive sampling methods can be found in [40] and [25].

---

**Algorithm A**: Adaptive sampling for GF with CoKriging and SERV

---

**Given:** Initial training set $\{x_{train}, y_{train}\} = \{(x_{LF1}, y_{LF1}), (x_{LF2}, y_{LF2}), ..., (x_{HF}, y_{HF})\}$, where it follows a nested data structure: $\{(x_{HF}, y_{HF})\} \subseteq \{(x_{LF_n}, y_{LF_n})\} \subseteq \cdots \subseteq \{(x_{LF1}, y_{LF1})\}$, HF model and LF model $f_i(x), i = HF, LF1, ... LF_n$, and the corresponding sampling cost $cost(i), i = HF, LF1, .. LF_n$

**Goal:** To improve the prediction accuracy of the HF surrogate model

**Define:** AF, stopping criteria

**Step 0:** Train CoKriging sequentially with $\{x_{train}, y_{train}\}$

**While** *not meeting stopping criteria* **do**

    1. First-stage optimization: Identify the location of interest $x_{next}$:

$$x_{next} = \arg\max AF(x_{HF})$$

    2. Second-stage optimization: Select infill sample with the maximum scaled expected variance reduction (SERV):

        **For** $i = LF1, LF2, ..., LF_n, HF$ **do**

        predict SERV of $x_{next}$ as $SERV_i = \frac{\rho_{l-1}\hat{s}_{Z_{l-1}}^2(x_{next}) + \hat{s}_{\delta_l}^2(x_{next})}{cost(i)}$, for $i > 2$

        **End**

    3. $l \leftarrow \arg\max(SERV_i)$

    4. Get infill sample $\{x_{next}, y_{next}\} \leftarrow$ Query $f_l(x)$ at $x_{next}$ to obtain $y_{next}$

        **For** $j = LF_{l-1}, ..., LF1$ **do**

        a. Query $f_j(x)$ at $x_{next}$ to obtain $y_{next}^j$

        b. Update infill set $\{x_{next}, y_{next}\} \leftarrow \{x_{next}, y_{next}\} \cup \{x_{next}, y_{next}^j\}$

        **End**

    5. Update training set $\{x_{train}, y_{train}\} \leftarrow \{x_{train}, y_{train}\} \cup \{x_{next}, y_{next}\}$

    6. Check if the stopping criteria is met.

    7. Refit CoKriging with $\{x_{train}, y_{train}\}$

**End**

**Output:** Updated training set, improved MF surrogate model

---

## A.3 BO Benchmark Method: Multi-fidelity Cost-aware Bayesian Optimization (MFCA)

We used the MFCA as our benchmark problem, which can be found in [41]. Note that we replaced the LMGP in the original framework with LVGP since they are mathematically identical in all the test cases.

---

**Algorithm B**: MFCA

---

**Given:** Initial multi-fidelity $D^k = \left\{ (\boldsymbol{x}^i, y^i) \right\}_{i=1}^{k}$, black-box functions $f(\boldsymbol{x}; j)$ and their corresponding sampling costs $O(j)$ where $j = [1, \dots, ds]$

**Goal:** Optimizing high-fidelity function (source $l \in [1, \dots, ds]$)

**Define:** Utility functions and stop conditions

**Step 0:** Train an LVGP and exclude highly biased low-fidelity sources based on its fidelity manifold.

**While** *stop conditions not met* **do**

1. Train an LVGP using $D^k$
2. Define the multi-fidelity cost-aware acquisition function:

   $\alpha_{LF}(\boldsymbol{x}; j) = \sigma_j(x)\phi(\frac{y_j^* - \mu_j(x)}{\sigma_j(x)})$, the exploration part of EI.

   $\alpha_{HF}(\boldsymbol{x}; l) = \mu_l(x) - y_l^*$, the improvement.

   $$\alpha_{MFCA}(\boldsymbol{x}; j) = \begin{cases} \dfrac{\alpha_{LF}(\boldsymbol{x}; j)}{O(j)}, & j \in \{1, \dots, ds\} \\ \dfrac{\alpha_{HF}(\boldsymbol{x}; l)}{O(l)}, & j = 1 \end{cases}$$

3. Solve the auxiliary optimization problem:
   $$[\boldsymbol{x}^{k+1}, j^{k+1}] = \arg\max_{\boldsymbol{x}, j} \alpha_{MFCA}(\boldsymbol{x}; j)$$
4. Query $f(; j)$ at point $x^{k+1}$ from source $j^{k+1}$ to obtain $y^{k+1}$
5. Update data: $D^{k+1} \leftarrow D^k \cup (\boldsymbol{x}^{k+1}, y^{k+1})$
6. Update counter: $k \leftarrow k + 1$

**End**

**Output:** Updated data $D^k$, optimal solution of HF model, LVGP

---

## Code Availability

The LVGP package for R can be accessed on Cran. The Matlab and Python code for MuFASa is available upon request from the corresponding author.